\documentclass[twoside,10pt]{article}

\usepackage[preprint,hyperref,abbrvbib]{jmlr2e_pja} 

\usepackage{graphicx} 

\usepackage{multirow}
\usepackage{colortbl}
\usepackage[table,svgnames]{xcolor}

\usepackage{amsmath,amssymb,amsfonts} 

\definecolor{rev_color}{rgb}{0.6,0.0,0.0}

\definecolor{revB_color}{rgb}{0.9,0.1,0.1}

\newcommand{\mb}[1]{\mathbf{#1}}
\newcommand{\bsy}[1]{\boldsymbol{#1}}

\newcommand{\sminus}{\text{-}}

\definecolor{atz_table1}{rgb}{0.85, 0.85, 0.85}
\definecolor{atz_table2}{rgb}{0.8, 0.8, 0.8}
\definecolor{atz_table3}{rgb}{0.75, 0.75, 0.75}

\jmlrheading{x}{xxxx}{xx-xx}{x/xx}{xx/xx}{xxxxxx}{Ryan Lopez and Paul J. Atzberger}
\editor{}

\ShortHeadings{GD-VAEs: Geometric Dynamic Variational Autoencoders}{Lopez and Atzberger}
\firstpageno{1}

\begin{document}

\title{GD-VAEs: Geometric Dynamic Variational Autoencoders for Learning Nonlinear Dynamics 
and Dimension Reductions}

\author{\name Ryan Lopez  \\
       \addr Department of Physics\\
       College of Creative Studies (CCS) \\
       University of California Santa Barbara\\
       Santa Barbara, CA 93106, USA
       \AND
       \name Paul J.\ Atzberger {\email atzberg@gmail.com} \\
       {\addr Department of Mathematics\\
       Department of Mechanical Engineering \\
       University of California Santa Barbara\\
       Santa Barbara, CA 93106, USA} \\
       {\small \url{https://atzberger.org}}} 

\editor{...}

\maketitle

\begin{abstract}
We develop data-driven methods incorporating geometric and topological
information to learn parsimonious representations of nonlinear dynamics from
observations.  The approaches learn nonlinear state-space models of the
dynamics for general manifold latent spaces using training strategies related
to Variational Autoencoders (VAEs).  Our methods are referred to as Geometric
Dynamic (GD) Variational Autoencoders (GD-VAEs).  We learn encoders and
decoders for the system states and evolution based on deep neural network
architectures that include general Multilayer Perceptrons (MLPs), Convolutional
Neural Networks (CNNs), and other architectures.  Motivated by problems
arising in parameterized PDEs and physics, we investigate the performance of
our methods on tasks for learning reduced dimensional representations of the
nonlinear Burgers Equations, Constrained Mechanical Systems, and spatial fields
of Reaction-Diffusion Systems.  GD-VAEs provide methods that can be used
to obtain representations in manifold latent spaces
for diverse learning tasks involving dynamics.
\end{abstract}

\begin{keywords}
variational autoencoders, dimension reduction, dynamical systems, 
scientific machine learning
\end{keywords}

\section*{Introduction}
We develop data-driven approaches for learning predictive
models and representations from observations of dynamical processes.  We introduce learning
methods that incorporate inductive biases leveraging known prior topological
and geometric information.  This allows for using insights from techniques in
the qualitative analysis of dynamical systems and other domain
knowledge.  In practice, the observation data can include experimental
measurements, large-scale computational simulations, or solutions
of complex dynamical systems for which we seek reduced descriptions.  Learning
representations with prescribed target properties can be used to help enhance
the robustness of predictions, yield more interpretable results, or
provide further insights into the underlying mechanisms generating observed
behaviors.

A central challenge in the dynamical setting is to learn informative
representations facilitating stable predictions for multiple future states of
the system~\citep{Nelles_Book_Nonlinear_Sys_Identification_2013}. 
For this purpose, we develop probabilistic autoencoders that
incorporate noise-based regularizations and geometric priors to learn smooth
reduced dimensional representations for the observations.  We train our probabilistic autoencoders
building on the framework of Variational Autoencoders
(VAEs)~\citep{KingmaWellingVAE2014}.  To incorporate known topological and
geometric information, we develop learning methods allowing for general
manifold latent spaces.  This presents challenges to obtain neural networks to
represent mappings to the manifold geometry and for gradient-based training.  We address
these issues by developing mappings based on extrinsic geometric descriptions
and by developing custom back-propagation methods for passage of gradient
information through our manifold latent spaces.  We demonstrate these 
methods for learning representations and performing prediction 
for constrained mechanical systems, parameterized non-linear PDES, 
Burger's equation, and Reaction-Diffusion systems.  

The general problem of learning dynamical models from a time series of
observations has a long history spanning many fields
\citep{Nelles_Book_Nonlinear_Sys_Identification_2013,
Sjoeberg_Black_Box_Nonlinear_Dyn_1995,
Chiuso_System_Identification_Review_2019,
Hong_System_Identification_Review_2008}.  This includes the fields of dynamical systems
\citep{Kutz_Brunton_book_ch_ROMs_2019, Sjoeberg_Black_Box_Nonlinear_Dyn_1995,
Mallet_Coifman_Manifold_Learning_LVM_Dyn_Sys_2015,
Kutz_Lusch_Nonlinear_Embeddings_2018, Mezic_Koopman_Review_2013,
Ohlberger_Redcuced_Basis_Review_2016,Hesthaven2016,
Crutchfield_Dynamics_Symbolic_1987, DeVore_Reduced_Basis_Methods_Ch_2017},
control~\citep{Brunton_Kutz_SINDy_2016,
Nelles_Book_Nonlinear_Sys_Identification_2013,
Kutz_DNN_Time_Step_Constraints_2018,DMD_Schmid_2010}, 
statistics \citep{Archer_Variational_State_Space_2015,
Jordan_Nonlinear_Dynamics_Theory_2020, Ghahramani_EM_Nonlinear_Dynamics_1998},
and machine learning \citep{Chiuso_System_Identification_Review_2019,
Hong_System_Identification_Review_2008,
Carlberg_Lee_Nonlinear_Dynamics_AE_2020,
Karniadakis_Raissi_Hidden_Phys_PDEs_2018, Bertozzi_CDMD_2019,
Perdikaris_Phys_Informed_Gen_Models_2018,
LopezAtzberger2020}.  Many of the most successful and widely-used
approaches rely on assumptions on the model structure, most commonly, that a
time-invariant linear dynamical systems (LDS) provides a good local approximation or that the 
noise is Gaussian. 
These include the Kalman Filter and extensions~\citep{Kalman1960, DelMoral1997,
Godsill2019, VanDerMerwe_Kalman_Unscented_2000, Wan_Kalman_Nonlinear_2000},
Proper Orthogonal Decomposition
(POD)~\citep{POD_Intro_Chatterjee_2000,Mendez2018}, and more recently Dynamic
Mode Decomposition (DMD) \citep{DMD_Schmid_2010, DMD_Kutz_Brunton_Book_2016,
DMD_Theory_and_App_Kutz_2014} and Koopman Operator approaches
\citep{Mezic_Koopman_Review_2013,Das2019,
Putinar_Mezic_cont_spectrum_Koopman_2020}.  
Methods for
learning nonlinear dynamics include the NARX and NOE approaches with function
approximators based on neural networks and other models classes
\citep{Nelles_Book_Nonlinear_Sys_Identification_2013,
Sjoeberg_Black_Box_Nonlinear_Dyn_1995}, 
sparse symbolic dictionary methods 
that are linear-in-parameters using LASSO~\cite{santosa1986linear,tibshirani1996regression} 
such as SINDy~\citep{Brunton_Kutz_SINDy_2016,
Schmidt2009, Sjoeberg_Black_Box_Nonlinear_Dyn_1995}, 
and dynamic Bayesian networks (DBNs), such as Hidden Markov Chains (HMMs) and
Hidden-Physics Models ~\citep{Karniadakis_Raissi_Hidden_Phys_PDEs_2018,
Pawar2020,Saul2020,Baum_HMM_1966,Krishnan_GMS_2017,
Ghahramani_EM_Nonlinear_Dynamics_1998}.

Many variants of autoencoders have been developed for making predictions of
sequential data.  This includes those based on Recurrent Neural Networks (RNNs) with
LSTMs and GRUs~\citep{Schmidhuber_LSTM_1997,
Goodfellow2016,Cho_Bengio_GRU_2014}.  
Approaches for incorporating topological information into latent
variable representations include the early works by Kohonen on Self-Organizing
Maps (SOMs) \citep{Kohonen_Original_Paper_Self_Organizing_Maps_1982} and Bishop
on Generative Topographical Maps (GTMs) based on density networks providing a
generative approach \citep{Bishop_Early_Generative_Topographical_Map_1996}.
While general RNNs provide a rich approximation
class for sequential data, they pose for dynamical systems challenges for
interpretability and for training to obtain predictions stable over many steps.
Autoencoders have also
been combined with symbolic dictionary learning for latent dynamics in
\citep{Kutz_Discovery_Coordinates_2019} providing some advantages for
interpretability and robustness.  Dictionary methods however require specification 
in advance of sufficiently expressive libraries of functions.  
Neural networks incorporating physical information have also been developed,
where some of the methods use regularizations introduced during training to
enhance stability
\citep{Carlberg_RNN_Dynamics_2020,Atzberger_ML_Discussion_Foundations_2018,
TraskAtzberger_GMLS_Nets_AAAI_2020,stinis2024sdyn,
Carlberg_Lee_Nonlinear_Dynamics_AE_2020,TraskAtzberger_GMLS_Nets_AAAI_2020,
Atzberger_mlmod_preprint_2021,
Erichson_AE_Lyapunov_Stable_Flow_2019}.  The work of
\citep{Bengio_Recurrent_LVM_2015} considers methods for processing of speech
and handwriting by investigating RNNs combined with VAEs to obtain 
more robust sequential models.

In our work, we learn dynamical models building on the VAE framework to train
probabilistic encoders and decoders between general manifold
latent spaces.  This provides additional regularizations and constraints 
to help promote parsimoniousness, disentanglement of features, robustness, 
and interpretability.  Prior VAE methods used for dynamical systems include
\citep{Hernandez_VAE_Dynamics_2018, Pearce_Latent_Dynamics_GP_2020,LopezAtzberger2020,
Girin_Dynamical_VAE_Review_2020, Chen_VAE_Dynamic_Motion_Primitives_2016,
Pearce_Latent_Dynamics_GP_2020,
Roeder_VAE_Dynamics_Hierarchical_2019}.  These works use primarily Euclidean
latent spaces and consider applications such as human motion capture and ODE
systems.  More recently, VAE methods using non-Euclidean latent spaces include
\citep{Jensen_Manifold_Latent_Model_2020, Kalatzis_VAE_B_Motion_Priors_2020, 
Chen_VAE_Learn_Flat_Manifolds_2020, 
LopezAtzberger2020,
Arvanitidis_NonEuclidean_Latent_Space_2018}. 
These incorporate the role of geometry by augmenting the prior
distribution
on a Euclidean latent space to bias encodings toward a manifold.
In the recent works \citep{Rey_Diffusion_VAE_2020,Kipf_VAE_Hyperspherical_2018,
Falorsi_Homeomorphic_VAE_2018}, explicit projection
procedures are introduced to map analytically or to sample through random walks
an embedded manifold.  These works primarily consider geometries based
on $SO(3)$, spheres, tori, and cylinders.  We develop in our work 
more explicit procedures that can be used with back-propagation 
for more general geometries, including
non-orientable manifolds. 
We discuss related methods in our proceedings
paper~\citep{LopezAtzbergerAAAI2020}.
We also discuss related methods for handling
geometry using neural operator approaches in our recent
papers~\cite{quackenbush2024geometric,quackenbush2025transferable}
and stochastic systems in~\cite{stinis2024sdyn}.

We introduce here further methods for more general latent space
representations, including non-orientable manifolds, and applications to
parameterized PDEs, constrained mechanical systems, and reaction-diffusion
systems.  We introduce general methods for non-Euclidean latent spaces in
terms of point-cloud representations of the manifold along with local gradient
information that can be utilized within general back-propagation frameworks.
This allows for manifolds that have complex shapes 
or arise from other unsupervised learning approaches.

The paper is organized as follows. 
In Section~\ref{sec_gd_vae}, we formulate the probabilistic encoder-decoder
models and use the Evidence Lower Bound (ELBO) to derive and motivate the loss
functions and regularizations used for training.   In
Section~\ref{sec_manifold_latent_spaces}, we discuss the challenges in learning
with manifold latent spaces.  We also develop methods for mapping to general
manifolds and for performing back-propagation for gradient-based training.  In
Section~\ref{sec_Results}, we present results demonstrating the methods for
learning representations for constrained mechanical systems, parameterized
non-linear PDEs, Burgers equation, and Reaction-Diffusion systems.  To
investigate the role of the non-linear mappings and noise regularizations in our
approaches, we also make comparisons of our methods with analytic techniques 
and some other widely-used data-driven methods.  We also present additional derivations in
Appendix~\ref{sec_backprop} and~\ref{sec_role_geo_struc}.  We present an
additional result on how learned covariances can be used to help identify geometric
structures in Appendix~\ref{sec_extract_geo_from_var}.  The introduced GD-VAE
methods provide ways to incorporate topological and geometric information 
when learning representations for dynamical tasks.

\section{Learning Nonlinear Dynamics with Variational Autoencoders (VAEs)}
\label{sec_gd_vae}

A central challenge in the non-linear setting is to learn
from observations informative parsimonious representations of the 
dynamics. In practice, 
observation data can include experimental measurements, large-scale
computational simulations, or solutions to more 
complicated dynamical systems for which we seek reduced descriptions.  
Such representations can then be used to make predictions,
to perform simulations, or as part of optimization for designing 
controllers
\citep{Nelles_Book_Nonlinear_Sys_Identification_2013}.  

We develop data-driven approaches building on the Variational Autoencoder (VAE)
framework~\citep{KingmaWellingVAE2014}.  In contrast to standard 
autoencoders, which can result in scattered disconnected 
encodings, VAEs train probabilistic encoders 
and decoders where noise provides additional regularizations.  In VAEs
this promotes smoother dependence on inputs,
more connectedness between the encodings, and disentanglement 
of encoded components~\citep{KingmaWellingVAE2014}.  

\begin{figure}[ht]
\centerline{\includegraphics[width=0.99\columnwidth]{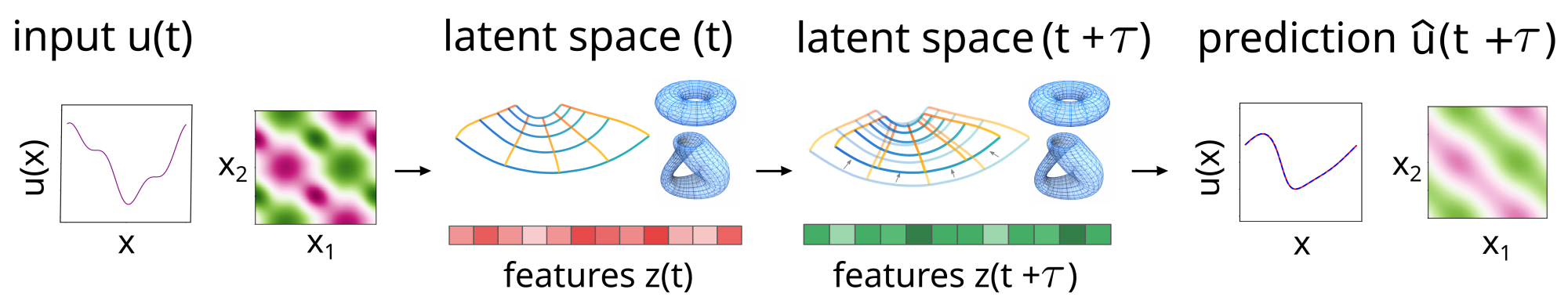}}
\caption{\textbf{Learning Nonlinear Dynamics with Encoders and Decoders.}  
Data-driven methods are developed for 
learning models to predict from $u(x,t)$ the non-linear evolution to
future states $u(x,t + \tau)$ for PDEs and other dynamical systems.  
Probabilistic autoencoders are utilized that learn representations 
$\mb{z}$ of $u(x,t)$ in reduced dimensional latent spaces with 
prescribed geometric and topological properties.  
The trainable models make predictions using learnable maps that (i) encode
an input $u(x,t) \in \mathcal{U}$ as $\mb{z}(t)$ in latent space \textit{(left)},
(ii) evolve the representation $\mb{z}(t) \rightarrow \mb{z}(t + \tau)$
\textit{(middle)}, (iii) decode the representation $\mb{z}(t + \tau)$ to
predict $\hat{u}(x,t + \tau)$ \textit{(right)}.
}
\label{fig:vae_learning_dynamics}
\end{figure}

\begin{figure}[ht]
\centerline{\includegraphics[width=0.8\columnwidth]{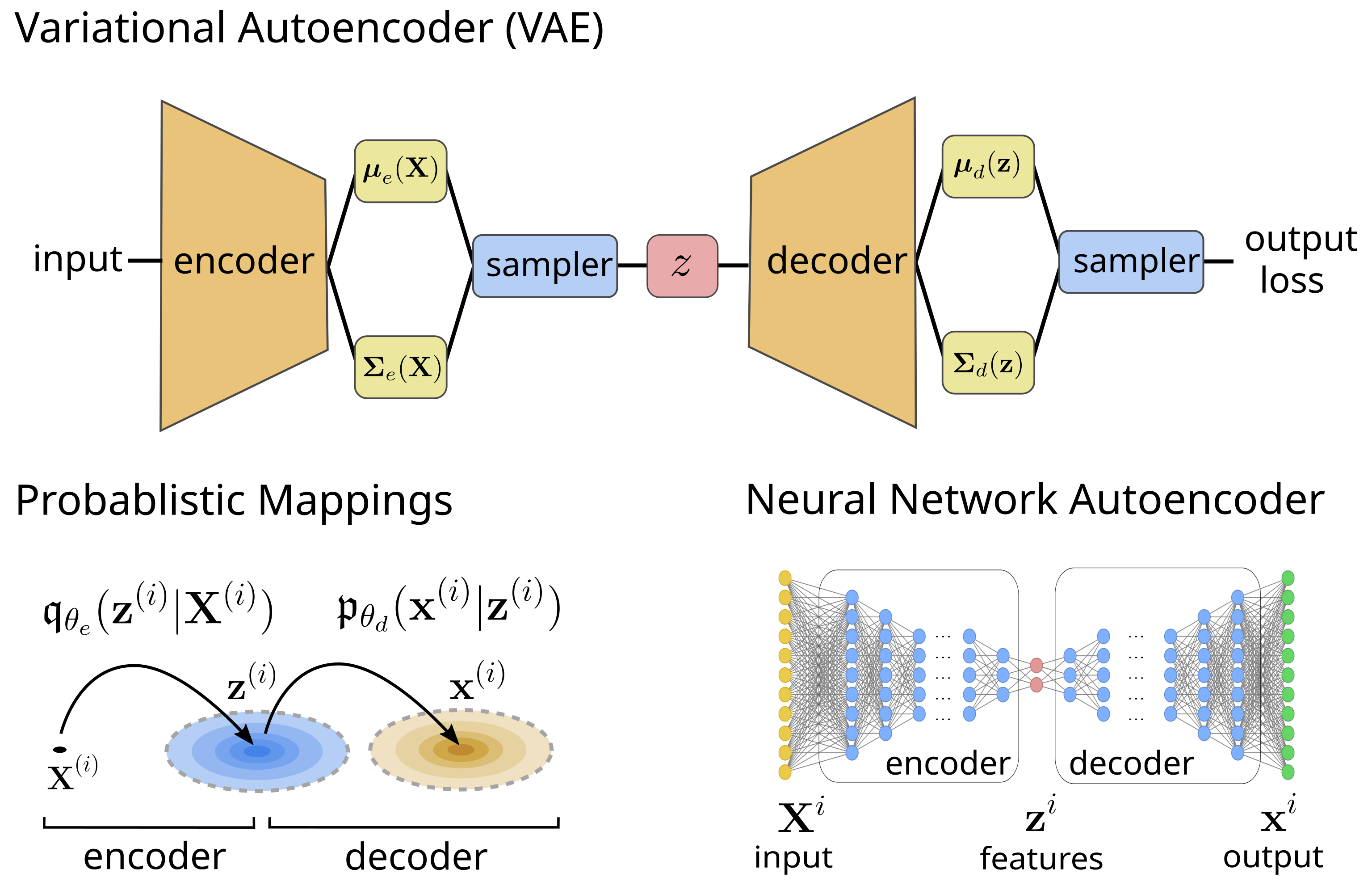}}
\caption{\textbf{Variational Autoencoder (VAE) Framework.}  
Representations 
of the non-linear dynamics are learned using probabilistic encoders and 
decoders trained using the VAE framework~\citep{KingmaWellingVAE2014}.
Our deep neural network (DNN) models have the VAE architecture 
shown at the \textit{(top)}.  This includes learnable 
probabilistic mappings, in contrast to more conventional
neural network autoencoders shown on the \textit{(bottom-right)}. 
GD-VAEs includes a probabilistic encoder with distribution $\mathfrak{q}_{\theta_e}$
and a probabilistic decoder with distribution  $\mathfrak{p}_{\theta_d}$,
shown on the \textit{(bottom-left)}. 
These DNNs are trained (i) to serve as feature extractors 
to represent functions $u(x,t)$ and their evolution in a reduced dimensional 
latent space as $\mb{z}(t) \rightarrow \mb{z}(t + \tau)$, 
and (ii) to serve as approximators that can 
construct predictions $u(x,t + \tau)$ using the features $\mb{z}(t+ \tau)$.
}
\label{VAE_schematic}
\end{figure}

We learn VAE predictors using a Maximum Likelihood Estimation (MLE)
approach.  For dynamic predictions $\mb{X} \rightarrow \mb{x}$, we use a Log
Likelihood (LL) which can be expressed using the factorization
$p_\theta(\mb{X},\mb{x}) = p_\theta(\mb{x}|\mb{X}))p_\theta(\mb{X})$ as
\begin{eqnarray}
\mathcal{J}_{LL} = \log(p_\theta(\mb{X},\mb{x})) =
\underbrace{\log(p_\theta(\mb{x}|\mb{X}))}_{\mathcal{J}_{LC}} +
\underbrace{\log(p_\theta(\mb{X}))}_{\mathcal{J}_{LX}}.
\end{eqnarray}
The $\mathcal{J}_{LC}$ gives the Log-Conditional (LC) Likelihood and
$\mathcal{J}_{LX}$ gives the marginal Log Likelihood of $\mb{X}$.  For
predicting the dynamics of $u(t)$, we discretize in time.  We sample
trajectories as $\{u(s_k)\}_{k = 0}^N$, where $s_k = t_0 + k\tau$ and $\tau$ is
the sampling time-scale.  
We learn representations for the discretized dynamics $\mb{X} \rightarrow
\mb{x}$ by letting $\mb{X} = u(s_k)$ and $\mb{x} = u(s_{k+1})$ with $p_\theta$
based on the dynamic autoencoder framework in
Figure~\ref{fig:vae_learning_dynamics} and~\ref{VAE_schematic}.  We use
variational inference to approximate the LL by the Evidence Lower Bound
(ELBO)~\citep{Blei_VI_Review_2017,KingmaWellingVAE2014}.  We use the concavity
of $\log(\cdot)$ and Jensen's Inequality for $\log \mathbb{E}[(\cdot)] \geq
\mathbb{E}[\log(\cdot)]$.  This yields the lower-bound

\begin{eqnarray}
\label{equ_lower_bound_LL}
\nonumber
\mathcal{J}_{LL} &=& \log(p(\mb{x}|\mb{X})p(\mb{X}))
= \log \mathbb{E}_{q(\mb{z}|\mb{X})}\left\lbrack 
\frac{p(\mb{x},\mb{z}|\mb{X})p(\mb{X})}{q(\mb{z}|\mb{X})} \right\rbrack 
= \log \mathbb{E}_{q(\mb{z}|\mb{X})}\left\lbrack 
\frac{p(\mb{x}|\mb{z})p(\mb{z}|\mb{X})p(\mb{X})}{q(\mb{z}|\mb{X})} \right\rbrack  \\
&\geq &
\mathbb{E}_{q(\mb{z}|\mb{X})}\left\lbrack  \log
\left(\frac{p(\mb{x}|\mb{z})p(\mb{z} | \mb{X})p(\mb{X})}{q(\mb{z}|\mb{X})}\right) 
\right\rbrack = 
\mathbb{E}_{q(\mb{z}|\mb{X})}\left\lbrack  \log
\left(\frac{p(\mb{x}|\mb{z})p(\mb{X}|\mb{z})p(\mb{z})}{q(\mb{z}|\mb{X})}\right) 
\right\rbrack
\\
\nonumber
&=& \underbrace{\mathbb{E}_{q(\mb{z}|\mb{X})}\left\lbrack  \log
\left(p(\mb{x}|\mb{z})\right) 
\right\rbrack}_{\mathcal{J}_{RE}} + 
\underbrace{\mathbb{E}_{q(\mb{z}|\mb{X})}\left\lbrack  \log
\left(p(\mb{X}|\mb{z})\right) 
\right\rbrack}_{\mathcal{J}_{RR}} + \underbrace{\mathbb{E}_{q(\mb{z}|\mb{X})}\left\lbrack  \log
\left(\frac{p(\mb{z})}{q(\mb{z}|\mb{X})}\right) 
\right\rbrack}_{\mathcal{J}_{KL}}, \\
\nonumber
 \Rightarrow \; \mathcal{J}_{LL} & \geq  &  \mathcal{J}_{RE} + 
 \mathcal{J}_{RR} + \mathcal{J}_{KL}.
\end{eqnarray}
Maximizing this lower bound corresponds to minimizing the loss 
$\mathcal{L}^B =  -\mathcal{J}_{RE}- \mathcal{J}_{RR}-\mathcal{J}_{KL}$.
We treat the $\mathcal{J}_{RR}$ and $\mathcal{J}_{KL}$ as regularization terms 
for the prediction and introduce additional adjustable 
parameters $\gamma,\beta$.  We use 
the regularized loss function
\begin{eqnarray}
\mathcal{L}^B =  -\mathcal{J}_{RE}- \gamma \mathcal{J}_{RR} - \beta \mathcal{J}_{KL}.
\end{eqnarray}

To simplify the notation for the expressions above, we have suppressed 
dependence on the distribution parameters $\theta$.  We assume here the considered
probability distributions factor to have an autoencoding-like property so that
$p(\mb{X},\mb{x},\mb{z}) := p(\mb{x} | \mb{z}) p(\mb{z} | \mb{X}) p(\mb{X})$.
This ensures the encoding $\mb{z}$ captures all the relevant information to
generate $\mb{x}$, so that $p(\mb{x}|\mb{z},\mb{X}) = p(\mb{x}|\mb{z})$.  The
$q(\mb{x}|\mb{z})$ gives the distribution for a probabilistic encoder map from
$\mb{x}$ to $\mb{z}$.  The $p(\mb{x}|\mb{z})$ gives a probabilistic decoder map
that 
reconstructs $\mb{x}$ from $\mb{z}$.  

We remark that the dynamic VAE can also be viewed abstractly as an 
autoencoder for reconstructing $(\mb{X},\mb{x})$ with constraints on the
distributions for encoding and decoding.  In particular, the factorization
$p(\mb{X},\mb{x},\mb{z}) = p(\mb{x} | \mb{z}) p(\mb{z} | \mb{X}) p(\mb{X})$ 
corresponds to $\mb{z}$ making $\mb{X}$ and $\mb{x}$ independent as a
data processing Markov chain $\mb{X} \rightarrow \mb{z} \rightarrow \mb{x}$.
The encoder distribution is constrained only to have a dependence on $\mb{X}$
as $q(\mb{z}|\mb{X},\mb{x}) = q(\mb{z}|\mb{X})$.  In the limit 
$\gamma \rightarrow 0$, which would remove the initial condition reconstruction
regularization $\mathcal{J}_{RR}$, the $\mb{X}$ would be encoded 
only to reconstruct $\mb{x}$.  We discuss the consequences of this and other
parameter choices in our studies below.

The encoding and decoding distributions in the variational approximations
will be taken to be learnable Gaussians of the form $p_\theta(\mb{x}|\mb{z}) =
\mathcal{N}(\mb{x};\bsy{\mu}_d(\mb{z};\theta),\bsy{\Sigma}_d(\theta))$ 
and $q_\theta(\mb{z}|\mb{X}) =
\mathcal{N}(\mb{z};\bsy{\mu}_e(\mb{X};\theta),\bsy{\Sigma}_e(\theta))$.  The
$\mathcal{N}$ denotes the probability density of the multi-variate Gaussian
with mean $\bsy{\mu}(\mb{z};\theta)$ and covariance $\bsy{\Sigma}(\theta)$.
In the case of no noise $\bsy{\Sigma}_e = 0$, the 
$\bsy{\mu}_e(\mb{X})$ would serve as a deterministic encoding map 
of $\mb{x}$ to obtain $\mb{z}$.  When there is noise $\bsy{\Sigma}_e \neq 0$, 
$\bsy{\mu}_e$ becomes the center location of the Gaussian in latent space 
and is the most probable encoding of 
$\mb{x}$ to obtain $\mb{z}$, see
Figure~\ref{VAE_schematic} \textit{(lower-left)}.     This holds similarly for
the probabilistic decoder map given by $\bsy{\mu}_d$ and $\bsy{\Sigma}_d$.  For
the decoder, the log probability gives a weighted $\ell^2$-norm penalizing
errors in reconstructing $\mb{x}$, since for a Gaussian
$-\log(p(\mb{x}|\mb{z})) =
\frac{1}{2}\| \mb{x} - \bsy{\mu}(\mb{z})  \|_{\Sigma}^2 + c_0$.  Here, the $c_0$ is
a constant independent of $\mb{x}$ and \begin{equation} \| \mb{x} -
\bsy{\mu}(\mb{z})  \|_{\Sigma}^2 = (\mb{x} - \bsy{\mu}(\mb{z}))^T \Sigma^{-1}
(\mb{x} - \bsy{\mu}(\mb{z})).  \end{equation} This holds similarly for
$-\log(p(\mb{X}|\mb{z}))$ in reconstructing $\mb{X}$ from $\mb{z}$. 

As mentioned above, we reformulate the maximization of the 
lower-bound in equation~\ref{equ_lower_bound_LL} as a minimization 
of the regularized loss $\mathcal{L}^B$.  The optimization then
can be interpreted as minimizing the reconstruction loss for the predicted future 
state $\mb{x}$ subject to additional regularizations
\begin{eqnarray}
\mathcal{L}_{RE} &=& -\mathcal{J}_{RE} = -\mathbb{E}_{q(\mb{z}|\mb{X})}\left\lbrack  \log
\left(p(\mb{x}|\mb{z})\right) 
\right\rbrack.
\end{eqnarray}
This includes regularization by 
reconstruction of the initial conditions $\mb{X}$ 
\begin{eqnarray}
\mathcal{L}_{RR} &=& -\mathcal{J}_{RR} = -\mathbb{E}_{q(\mb{z}|\mb{X})}\left\lbrack  \log
\left(p(\mb{X}|\mb{z})\right)
\right\rbrack,
\end{eqnarray}
and regularization based on 
reducing the Kullback-Leiber
(KL)-Divergence between $q(\mb{z}|\mb{X})$ and the prior distribution $p(\mb{z})$,
\begin{equation}
\mathcal{L}_{KL} = -\mathcal{J}_{KL} = -\mathbb{E}_{q(\mb{z}|\mb{X})}
\left\lbrack  \log
\left(\frac{p(\mb{z})}{q(\mb{z}|\mb{X})}\right) 
\right\rbrack = \mathcal{D}_{KL}\left(q(\mb{z}|\mb{X}) | p(\mb{z}))  \right).
\end{equation}

We also split the latent codes as $\bar{\mb{z}} = (\mb{z},\mb{z}')$ and
introduce the additional distribution $p_{\theta_\ell}(\mb{z}' | \mb{z})$ which
models steps of the latent space dynamics $\mb{z}(t) \rightarrow \mb{z}(t +
\tau)$ where $\mb{z} = \mb{z}(t)$ and $\mb{z}'=\mb{z}(t + \tau)$.  We restrict
the decoder distributions to have dependence  $p(\mb{X}|\bar{\mb{z}}) =
p(\mb{X}|\mb{z})$ and $p(\mb{x}|\bar{\mb{z}}) = p(\mb{x}|\mb{z}')$.  In the
case of deterministic dynamics, we alternatively can express this by
introducing the latent-space mapping $\mb{z}' = f_{\theta_\ell}(\mb{z})$
parameterized by $\theta_\ell$.  We also augment the $\mathcal{D}_{KL}$ term to
serve as an adjustable regularization weighted by the parameter $\beta$ and
also allow for adjusting the relative strength of the reconstructions for
$\mb{x}$ and $\mb{X}$ using the parameter $\gamma$.  Our models have the learnable
parameters $\theta=(\theta_e,\theta_d,\theta_\ell)$ with $\theta_e$ for the
encoder, $\theta_d$ for the decoder, and $\theta_\ell$ for the dynamics. 

In summary, we train our models based on minimizing the loss function
$\mathcal{L}^B = \sum_{i=1}^m \mathcal{L}^B_i$ with

\begin{eqnarray}
\label{equ_vae}
\theta^* &=& \arg\min_{\theta \in \Theta}\; \mathcal{L}^B(\theta_e,
\theta_d,\theta_\ell;\{\mb{X}^{(i)},\mb{x}^{(i)}\}_{i=1}^m), \;\;\;
\label{equ:vae_loss}
\mathcal{L}_i^{B} =  \mathcal{L}_{RE}^{(i)} + \mathcal{L}_{KL}^{(i)} 
+ \mathcal{L}_{RR}^{(i)}, \\
\nonumber
\mathcal{L}_{RE}^{(i)} &=&
-E_{\mathfrak{q}_{\theta_e}(\mb{z}|\mb{X}^{(i)})}\left\lbrack 
\log \mathfrak{p}_{\theta_d}(\mb{x}^{(i)} | \mb{z}') \right\rbrack, \;\;
\nonumber
\mathcal{L}_{KL}^{(i)} =
\beta\mathcal{D}_{KL}\left(\mathfrak{q}_{\theta_e}(\mb{z}|\mb{X}^{(i)}) 
\, \| \, 
\tilde{\mathfrak{p}}_{\theta_p}(\mb{z})\right), \\
\nonumber
\mathcal{L}_{RR}^{(i)} &=&
-\gamma 
E_{\mathfrak{q}_{\theta_e}(\mb{z}|\mb{X}^{(i)})}\left\lbrack 
\log \mathfrak{p}_{\theta_d}(\mb{X}^{(i)} | \mb{z}) \right\rbrack.
\end{eqnarray}

The distribution for the
probabilistic encoding map
is denoted by $\mathfrak{q}_{\theta_e}$, the distribution
of the probabilistic decoding map by $\mathfrak{p}_{\theta_d}$, 
and the  
prior distribution by $\tilde{\mathfrak{p}}_{\theta_d}$.  The 
prior $p(\mb{z}) = \tilde{\mathfrak{p}}_{\theta_p}$ is 
used for regularizing the organization of the latent-space
representation $\mb{z}$.  Throughout, we use a Gaussian 
prior of the form $\tilde{\mathfrak{p}}_{\theta_p}(\mb{z}) = 
\mathcal{N}(\mb{z};\bsy{\mu}_p,\bsy{\Sigma}_p)$
with mean zero $\bsy{\mu}_p = 0$ and 
diagonal covariance $\bsy{\Sigma}_p = \sigma_p^2 \mathcal{I}$. 
We remark other priors also can be considered that are further tailored to the task.
The
model is trained using batches of $m$
observation samples $\{(\mb{X}^{(i)},\mb{x}^{(i)})\}_{i=1}^m$.  
For a dynamical
system with a collection of sampled trajectories $\{u^k(s)\}_k$, 
we have that $\mb{X}^{(i)}$ samples the
initial state $u^{k_i}(t_i)$ and $\mb{x}^{(i)}$ samples the future state $u^{k_i}(t_i +
\tau)$.  The $\tau$ gives the time-scale over which the prediction is made.
The $k_i$ gives the trajectory index $k$ and the $t_i$ the 
sampled initial time $t$ from the trajectory.

The loss function in equation~\ref{equ:vae_loss} provides a regularized form of
MLE.  The three
terms comprising $\mathcal{L}^B$ can be interpreted as follows
(i) $\mathcal{L}_{RR}$ for the log likelihood of
reconstructing samples, (ii) $\mathcal{L}_{RE}$ is the log likelihood of
correctly predicting samples after a single time step, and  (iii)
$\mathcal{L}_{KL}$ is a regularization term based on the Kullback-Leibler (KL)
Divergence between the encoding distribution and latent space prior
distribution.

For the trainable encoder and decoder probability distributions,
we use parameterizations for the Gaussians of the form 
\begin{eqnarray}
\mathfrak{q}_{\theta_e}(\mb{z}|\mb{X}) &=&
\mathcal{N}(\mb{z};\bsy{\mu}_e(\mb{X};\theta_e),\bsy{\Sigma}_e(\mb{X};\theta_e))
\\
\mathfrak{p}_{\theta_d}(\mb{X}|\mb{z}) &=&
\mathcal{N}(\mb{X};\bsy{\mu}_d(\mb{z};\theta_d),\bsy{\Sigma}_d(\mb{z};\theta_d))
\\
\mathfrak{p}_{\theta_d}(\mb{x}|\mb{z}') &=&
\mathcal{N}(\mb{x};\bsy{\mu'}_d(\mb{z}';\theta_d),\bsy{\Sigma}_d'(\mb{z}';\theta_d)).
\end{eqnarray}
The $\mathcal{N}$ denotes the density of the multivariate Gaussian with mean
$\bsy{\mu}$ and covariance $\bsy{\Sigma}$.  We also can express this more 
directly in terms of the probabilistic maps and resulting random variables as 
\begin{eqnarray}
\begin{array}{llllll}
\label{equ_prob_mappings}
\mb{z}  &\sim&  \mathfrak{q}_{\theta_e}(\mb{z}|\mb{X}), 
& 
\mb{z} &=& \bsy{\mu}_e(\mb{X};\theta_e) + \bsy{\eta}_e\\
\mb{z}' &\sim&  \mathfrak{p}_{\theta_\ell}(\mb{z}'|\mb{z}),  
&
\mb{z}' &=&  f_{\theta_\ell}(\mb{z})\\
\mb{x}  &\sim&  \mathfrak{p}_{\theta_d}(\mb{x}|\mb{z}'), 
&
\mb{x} &=& \bsy{\mu}'_d(\mb{z}';\theta_d) + \bsy{\eta}_d' \\
\mb{X}  &\sim&  \mathfrak{p}_{\theta_d}(\mb{X}|\mb{z}), 
&
\mb{X} &=& \bsy{\mu}_d(\mb{z};\theta_d) + \bsy{\eta}_d.
\end{array}
\end{eqnarray}
The $\bsy{\eta}_*$ denote noise regularizations provided
by Gaussian random variables with mean zero and covariance $\bsy{\Sigma}_*$.

The terms that can be learned in this dynamic VAE framework are
$(\bsy{\mu}_e,\bsy{\Sigma}_e,\bsy{\mu}_d,\bsy{\Sigma}_d,
\bsy{\mu}_d',\bsy{\Sigma}_d',f_{\theta_\ell})$,
which are parameterized by $\theta = (\theta_e,\theta_d,\theta_\ell)$. 
The $\bsy{\mu}_e,\bsy{\mu}_d,\bsy{\mu}_d'$ denote learnable functions encoding
or decoding the inputs.  These can be represented using deep neural networks or
other model classes, see Figure~\ref{VAE_schematic}.
The noise serves as an additional source of regularization to help promote
smoothness and connectedness of the encodings, 
see Figure~\ref{VAE_schematic} \textit{(lower-left)}.
In practice, while the
variances are learnable, for many problems it can be useful to treat the
$\bsy{\Sigma}_*$ as hyper-parameters.    We discuss some ways to use 
learnable covariances $\bsy{\Sigma}_*$ to extract from the data useful geometric 
structures in Appendix~\ref{sec_extract_geo_from_var}.

We learn predictors for the dynamics by training over samples of evolution pairs
$\{(u_i^{k_i},u_{i+1}^{k_i})\}_{i=1}^m$, where $i$ denotes the sample index and $u_i^{k_i}
= u^{k_i}(s_i)$ with $s_i = t_0 + i\tau$ for a time-scale $\tau$.  To make
predictions, the learned models use the following stages: (i) extract from
$u(t)$ the features $z(t)$, (ii) evolve $z(t) \rightarrow z(t + \tau)$, (iii)
predict using $z(t + \tau)$ the $\hat{u}(t + \tau)$, see 
Figures~\ref{fig:vae_learning_dynamics} and~\ref{VAE_schematic}.
By function composition, the learned model also can 
make multi-step predictions for the dynamics.

\section{Learning with Manifold Latent Spaces having General Geometries and
Topologies}
\label{sec_manifold_latent_spaces}

\begin{figure}[ht]
\centerline{\includegraphics[width=0.8\columnwidth]{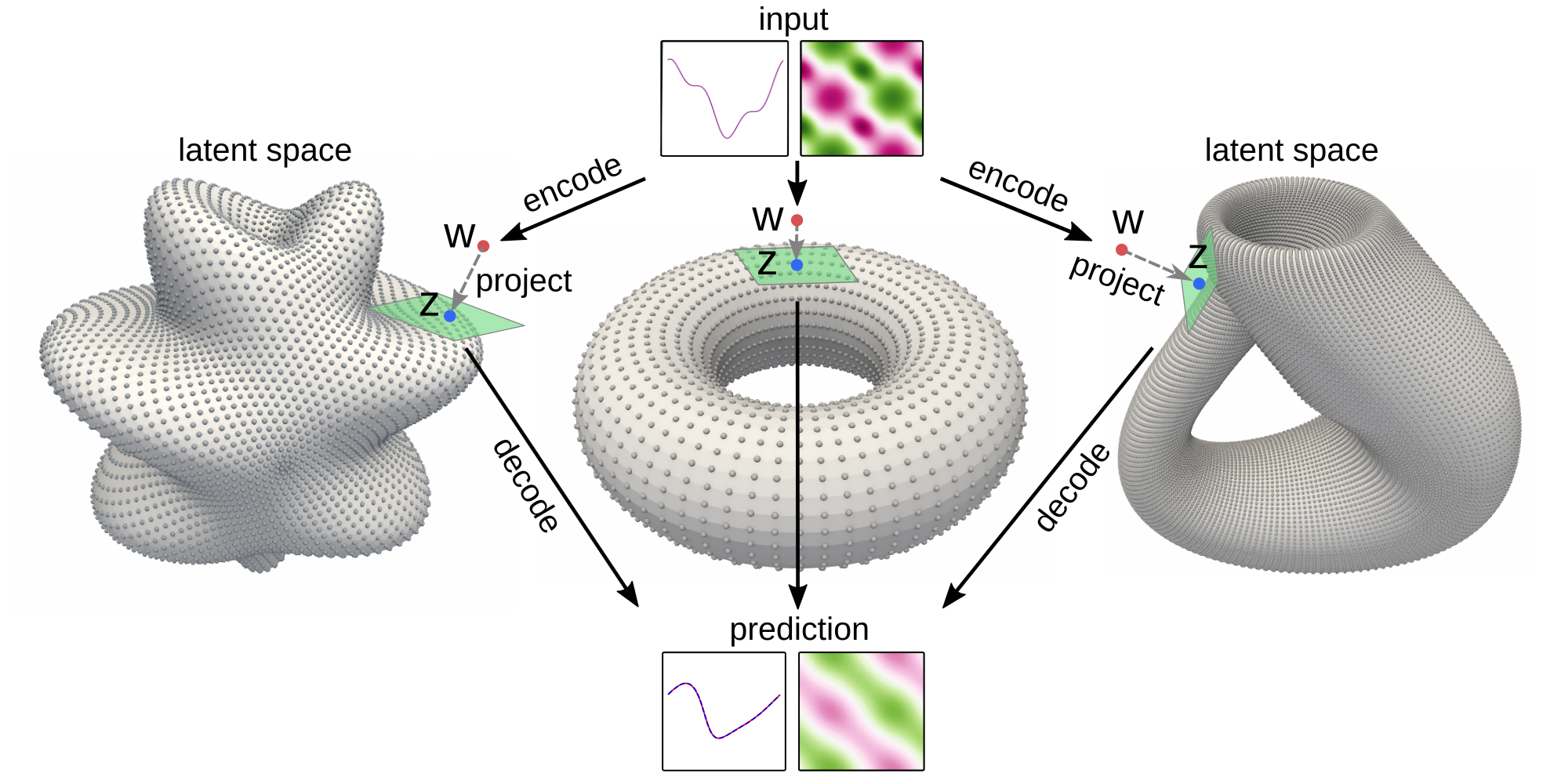}}
\caption{\textbf{Learnable Mappings for Manifold Latent Spaces.}  We develop
methods for using manifold latent space representations having general
geometries and topologies through projection maps.  For inter-operability with
widely used trainable mappings, such as neural networks, we use a strategy of
mapping inputs first to a point $\mb{w} =
\tilde{\mathcal{E}}_\phi(\mb{x})\in\mathbb{R}^N$ for the embedding space which
is then projected to a point in the manifold $\mb{z} = \Lambda(\mb{w}) \in
\mathcal{M}\subset \mathbb{R}^N$.  This provides trainable mappings for general
manifold latent spaces.  In practice, we can represent the manifold and compute
projections based on general point cloud representations, analytic
descriptions, product spaces, or other approaches.}
\label{fig:manifold_map_embed}
\end{figure}

For many systems, parsimonious representations can be obtained by working with
non-Euclidean manifold latent spaces.  This includes tori for doubly periodic
systems or even non-orientable manifolds, such as a klein bottle as arises in
imaging and perception studies~\citep{Carlsson2008}.  For this purpose, we
learn encoders $\mathcal{E}$ over a family of mappings to a prescribed manifold
$\mathcal{M}$ of the form
\begin{eqnarray}
\mb{z} = \mathcal{E}_\phi(\mb{x}) 
= \Lambda(\tilde{\mathcal{E}}_{\phi}(\mb{x})) 
= \Lambda(\mb{w}), \; \mbox{where} \; \; \mb{w} 
= \tilde{\mathcal{E}}_{\phi}(\mb{x}).
\label{equ_manifold_map}
\end{eqnarray}
The $\mathcal{E}_\phi$ denotes a candidate encoder to the manifold when the 
parameters are $\phi$.  This can be used as part of the encoder map with  
$\bsy{\mu}_e(\cdot) = \mathcal{E}_\phi(\cdot)$, see equation~\ref{equ_prob_mappings}
and Figure~\ref{fig:manifold_map_embed}.

To generate a family of maps over which we can learn in practice, we use that a
smooth closed manifold $\mathcal{M}$ of dimension $m$ can be embedded within
$\mathbb{R}^{2m}$, as supported by the Whitney Embedding Theorem
\citep{Whitney1944}.  We obtain a family of maps to the manifold by constructing
maps in two steps using equation~\ref{equ_manifold_map}.  
In the first step, we use an unconstrained encoder 
$\tilde{\mathcal{E}}$ from $\mb{x}$ to a point $\mb{w}$ in the embedding space. 
In the second step, we use a map $\Lambda$ that projects a point $\mb{w} \in
\mathbb{R}^{2m}$ to a point $\mb{z} \in \mathcal{M} \subset
\mathbb{R}^{2m}$ within the embedded manifold.  In this way, 
$\tilde{\mathcal{E}}$ can be any learnable mapping from $\mathbb{R}^n$ to 
$\mathbb{R}^{2m}$ for which there are many model classes including 
neural networks.  To obtain a particular manifold map, the $\tilde{\mathcal{E}}$  
only needs to learn an equivalent mapping from 
$\mb{x}$ to $\mb{w}$, where $\mb{w}$ is in the appropriate equivalence class
$\mathcal{Q}_{\mb{z}}$ of a target point $\mb{z}$ on the manifold, 
$\mb{w} \in \mathcal{Q}_{\mb{z}} = \{\mb{w} \; | \; \Lambda(\mb{w}) = \mb{z} \}$.  
Here, we accomplish this in practice two ways: (i) we provide an
analytic mapping $\Lambda$ to $\mathcal{M}$, (ii) we provide a high resolution
point-cloud representation of the target manifold along with local gradients
and use for $\Lambda$ a quantized or interpolated mapping to the nearest point on
$\mathcal{M}$.  We provide more details of our methods below and in Appendix A.  

In practice, we can view the projection map $\mb{z} = \Lambda(\mb{w})$ to the manifold 
as the solution of the optimization problem 
\begin{eqnarray}
\label{equ_z_ls}
\mb{z}^* = \arg\min_{\mb{z} \in \mathcal{M}} \frac{1}{2}\|\mb{w} -
\mb{z}\|_2^2. 
\end{eqnarray}
We can always express patches of a smooth manifold using local coordinate
charts $\mb{z} = \sigma^k(\mb{u})$ for $\mb{u} \in \mathcal{U} \subset
\mathbb{R}^{m}$.  For example, we could use in practice a local Monge-Gauge
quadratic fit to a point cloud representation of the manifold, as
in~\cite{Atzberger_GMLS_Surf_PDE_2019}.  We can express $\mb{z}^* =
\sigma^{k^*}(\mb{u}^*)$ for some chart $k^*$ for solution of
equation~\ref{equ_z_ls}.  In terms of the collection of coordinate charts
$\{\mathcal{U}^k\}$ and local parameterizations $\{\sigma^{k}(\mb{u})\}$, we
can express this as 
\begin{eqnarray}
\label{equ_ls_Phi}
\mb{u}^*,k^* = \arg\min_{k,\mb{u} \in \mathcal{U}^k} \Phi_k(\mb{u},\mb{w}), 
\; \mbox{where}\;\; \Phi_k(\mb{u},\mb{w}) = \frac{1}{2}\|\mb{w} -
\sigma^k(\mb{u})\|_2^2.
\end{eqnarray} 
The $\mb{w}$ is the input and $\mb{u}^*,k^*$ is the solution of 
equation~\ref{equ_ls_Phi}.  This gives the coordinate-based representation 
$\mb{z}^* = \sigma^{k^*}(\mb{u}^*)=\Lambda(\mb{w})$.
For smooth parameterizations $\sigma(\mb{u})$, the optimal solutions $\mb{u}^*(\mb{w})$ 
satisfies the following implicit equation arising from the optimization problem,
\begin{eqnarray}
\label{equ_G_implicit}
G(\mb{u}^*,\mb{w}) := \nabla_{\mb{u}}
\Phi_{k^*}(\mb{u}^*,\mb{w}) = 0.  
\end{eqnarray} 
During learning with back-propagation, we need to be able to compute the 
contributions to the loss function of the gradient 
\begin{eqnarray}
\nabla_\phi \mb{z}^* = \nabla_\phi \sigma^k(\mb{u}^*) = 
\nabla_\phi \Lambda\left(\tilde{\mathcal{E}}_\phi(\mb{x})\right)
= \nabla_{\mb{w}} \Lambda(\mb{w}) \nabla_\phi \tilde{\mathcal{E}}_\phi,
\label{equ_z_star}
\end{eqnarray}
where $\mb{w} = \tilde{\mathcal{E}}_\phi $.  If we approach training models
using directly the expressions of equation~\ref{equ_z_star}, we would need ways
to compute both the gradients $\nabla_\phi \tilde{\mathcal{E}}_\phi$ and
$\nabla_{\mb{w}} \Lambda(\mb{w})$.  While the gradients $\nabla_\phi
\tilde{\mathcal{E}}_\phi$ can be obtained readily for many model classes, such
as neural networks when using back-propagation, the gradients $\nabla_{\mb{w}}
\Lambda(\mb{w})$ pose additional challenges.  If $\Lambda$ can be expressed
analytically, then back-propagation techniques can be employed
directly.  However, in practice, $\Lambda$ will often result from a numerical
solution of the optimization problem in equation~\ref{equ_z_ls} which uses
equation~\ref{equ_ls_Phi}.  We show how in this setting alternative approaches
can be used to obtain the contributions to the loss function of the 
gradient $\nabla_\phi \mb{z}^* =\nabla_\phi
\Lambda\left(\tilde{\mathcal{E}}_\phi(\mb{x})\right)$.

To account for the contributions of such terms to the loss, we instead derive 
expressions for the variations of $\mb{u}^*$ and $\mb{w}$ using that they are 
related by $G(u^*,\mb{w}) = 0$.  If we parameterize $\phi = \phi(\gamma)$
by some scalar parameter $\gamma$, then we have $G(\mb{u}^*(\gamma),\mb{w}(\gamma)) = 0$ 
for each $\gamma$.  For example, this is motivated by taking
$\mb{w}(\gamma) = \tilde{\mathcal{E}}_\phi(\mb{x}(\gamma))$ and $\phi =
\phi(\gamma)$ for some path $(\mb{x}(\gamma),\phi(\gamma))$ in the input and
parameter space $(\mb{x},\phi) \in \mathcal{X}\times \mathcal{P}$.  We can
obtain the needed gradients by determining the variations of $\mb{u}^* =
\mb{u}^*(\gamma)$ by using equation~\ref{equ_G_implicit}. 
This follows since
$\mb{z}^* = \sigma^k(\mb{u}^*)$ and $\nabla_\phi \mb{z}^* = \nabla_{\mb{u}}
\sigma^k(\mb{u}^*)\nabla_\phi \mb{u}^*$.  The $\nabla_{\mb{u}}
\sigma^k(\mb{u}^*)$ often can be readily obtained analytically or 
numerically from back-propagation.  The more challenging term to compute 
is $\nabla_\phi \mb{u}^*$.  This can be obtained from 
${d\mb{u}^*}/{d\gamma}$ by considering a sufficiently rich variety of 
paths in parameter space $\phi = \phi(\gamma)$ that probe each direction.  
From equation~\ref{equ_G_implicit}, 
this allows us to express
the gradients using the Implicit Function Theorem as 
\begin{eqnarray}
\label{equ_grad_G_zero}
0 = \frac{d}{d\gamma} 
G(\mb{u}^*(\gamma),\mb{w}(\gamma))
= \nabla_{\mb{u}} G \frac{d\mb{u}^*}{d\gamma} 
+ 
\nabla_{\mb{w}} G \frac{d\mb{w}}{d\gamma}.
\end{eqnarray}
The term typically posing the most significant computational challenge 
is ${d\mb{u}^*}/{d\gamma}$ since $\mb{u}^*$ is
obtained numerically from the optimization problem in
equation~\ref{equ_ls_Phi}.  We solve for it using the expression in
equation~\ref{equ_grad_G_zero} to obtain
\begin{eqnarray}
\label{equ_ls_adjoint}
\frac{d\mb{u}^*}{d\gamma} = -\left[\nabla_{\mb{u}} G \right]^{-1}
\nabla_{\mb{w}} G \frac{d\mb{w}}{d\gamma}.
\end{eqnarray}
This only requires that we can evaluate for a given $(\mb{u},\mb{w})$ the local
gradients $\nabla_{\mb{u}} G$, $\nabla_{\mb{w}} G$, $d\mb{w}/d\gamma$, and that
$\nabla_{\mb{u}} G$ is invertible. Computationally, this only requires us to
find numerically the solution $\mb{u}^*$ and evaluate numerically the expression
in equation~\ref{equ_ls_adjoint} for a given $(\mb{u}^*,\mb{w})$.  This allows
us to avoid needing to compute directly
$\nabla_{\mb{w}}\Lambda_\mb{w}(\mb{w})$. Now since 
$\nabla_\phi \mb{z}^* = \nabla_{\mb{u}}\sigma^k(\mb{u}^*)\nabla_\phi \mb{u}^*$
and since we can compute $\nabla_\phi \mb{u}^*$ using equation~\ref{equ_ls_adjoint},
we can numerically evaluate $\nabla_\phi \mb{z}^*$. 
This provides the needed gradients for training our models with  
manifold latent spaces.

For learning via back-propagation, we use these results to assemble the needed
gradients.  For our manifold encoder maps $\mathcal{E}_\theta =
\Lambda(\tilde{\mathcal{E}}_\theta(\mb{x}))$, we use the following approach.
Using $\mb{w} = \tilde{\mathcal{E}}_\theta(\mb{x})$, 
we first find numerically the closest point in the manifold $\mb{z}^* \in
\mathcal{M}$ and represent it as $\mb{z}^* = \sigma(\mb{u}^*) = \sigma^{k^*}(\mb{u}^*)$ for
some chart $k^*$.  Next, using this chart we compute the gradients using that 
\begin{eqnarray}
\label{equ_G_grad_detail1}
G = \nabla_{\mb{u}} \Phi(\mb{u},\mb{w}) 
= -(\mb{w} - \sigma(\mb{u}))^T\nabla_{\mb{u}} \sigma(\mb{u}).
\end{eqnarray}
We take in equation~\ref{equ_G_grad_detail1} a column vector convention with
$\nabla_{\mb{u}} \sigma(\mb{u}) = [\sigma_{u_1} | \ldots | \sigma_{u_k}]$.  
We next compute 
\begin{eqnarray}
\label{equ_G_next}
\nabla_{\mb{u}} G = \nabla_{\mb{u}\mb{u}} \Phi 
= \nabla_{\mb{u}}\sigma^T \nabla_{\mb{u}} \sigma 
- (\mb{w} -
\sigma(\mb{u}))^T\nabla_{\mb{u}\mb{u}} \sigma(\mb{u})
\end{eqnarray}
and 
\begin{eqnarray}
\label{equ_G_w_comp}
\nabla_{\mb{w}} G = \nabla_{\mb{w},\mb{u}} \Phi = -I \nabla_{\mb{u}} \sigma(\mb{u}).
\end{eqnarray}
From equations~\ref{equ_G_next} and~\ref{equ_G_w_comp},
the gradients $\nabla_{\mb{u}} G$, $\nabla_{\mb{w}} G$,
and using equation~\ref{equ_ls_adjoint}, we compute $\nabla_\phi \mb{z}^*$.
This allows us to learn VAEs with latent spaces for $\mb{z}$ with general
specified topologies and controllable geometric structures. 
We provide additional details in Appendix A. 
We refer to this learning framework for data-driven modeling of 
dynamics on latent spaces having general geometries as 
Geometric Dynamic Variational Autoencoders (GD-VAEs).

We remark that the geometry of manifolds can be represented in many different 
ways using either extrinsic or intrinsic descriptions.  The algorithms 
leverage extrinsic descriptions which allows for avoiding the need for
local coordinate charts and instead uses a global parameterization 
(over-parameterization) of the manifold.  While many of the numerical 
calculations are performed in $\mathbb{R}^n$, we should emphasize that 
the learning methods still gain the benefits of the lower dimensionality 
of the manifold and still gain the benefits of the constraints arising from 
the latent-spaces geometric and topological structure.  For instance,
the decoder only needs to learn correct mappings from the subset of points 
that lie within the manifold.  The manifold latent-space results in 
learning methods gaining statistical power from the training samples 
since their mappings to and from the manifold results overall in 
needing to learn responses over a smaller set of inputs.

The manifold latent spaces provided by GD-VAEs also 
can be combined with the use of tailored prior distributions
$\mathfrak{p}(\mb{z})$ for the manifold setting. 
For simplicity, we primarily use here generic Gaussian priors similar to those 
used on the ambient space. When restricted to the geometric structure this gives
a conditional probability distribution over the manifold.  Our aim here is 
to use the priors to help with identifiability by setting the length-scale of 
the embedding maps and to help with learning by driving training steps 
toward a common target encoding.  The GD-VAE methods
are flexible and any prior for which the KL-Divergence regularization can be
computed can be used in practice.  Priors can be developed for
specific manifold latent-spaces leveraging further knowledge 
and information relevant to a target machine learning task.

GD-VAEs provide ways to leverage topological and geometric
information to help unburden the encoder and decoder from having to learn
these embedding structures as part of training.  As we shall discuss
below in the context of specific examples, the manifold latent space 
structures also can help with reduced sensitivity to noise and obtaining more robust and
parsimonious representations.  This increased level of organizational structure 
also can be used to help with identifiability and with interpretability of the
learned representations.

\section{Results}
\label{sec_Results}
We report on the performance of the methods.  
We consider examples that
include learning the dynamics of the non-linear Burger's PDEs  
and reaction-diffusion PDEs.  We also investigate 
learning representations
for constrained mechanical systems.
We consider the roles of manifold latent spaces having
different dimensions, topology, and geometry.  

\subsection{Burgers' Equation of Fluid Mechanics: Learning Nonlinear PDE
Dynamics}
\label{sec_burgers_equ}

We consider the nonlinear viscous Burgers' equation
\begin{eqnarray}
\label{equ_burgers_equ}
u_t = -uu_x + \nu u_{xx},
\label{eqn:burgers}
\end{eqnarray}
where $\nu$ is the viscosity~\citep{Bateman1915,Hopf1950}.  We consider
periodic boundary conditions on $\Omega = [0,1]$.  Burgers equation is
motivated as a mechanistic model for the fluid mechanics of advective transport
and shocks, and serves as a widely used benchmark for analysis and
computational methods.

The nonlinear Cole-Hopf Transform $\mathcal{CH}$ can be used to relate Burgers
equation to the linear Diffusion equation $\phi_t = \nu
\phi_{xx}$~\citep{Hopf1950}.  This provides a representation of the solution
$u$ 
\begin{eqnarray}
\label{equ_cole_hopf}
\phi(x,t) & = & \mathcal{CH}[u] = \exp \left( -\frac{1}{2\nu}
\int_0^x u(x',t)dx'\right),\;\;\;\;
u(x,t)  =  \mathcal{CH}^{-1}[\phi] = -2\nu 
\frac{\partial}{\partial x} \ln \phi(x,t).
\end{eqnarray}
This can be represented by the Fourier expansion
\begin{eqnarray}
\nonumber
&& \phi(x,t) = \sum_{k=-\infty}^{\infty} \hat{\phi}_k(0) 
\exp(-4\pi^2 k^2 \nu t) \cdot \exp(i2\pi kx).
\end{eqnarray}
The $\hat{\phi}_k(0) = \mathcal{F}_k[\phi(x,0)]$ and $\phi(x,t) =
\mathcal{F}^{-1}[\{\hat{\phi}_k(0) \exp(-4\pi^2 k^2 \nu t)\}]$ with
$\mathcal{F}$ the Fourier transform. This provides an analytic representation
of the solution of the viscous Burgers equation $u(x,t) =
\mathcal{CH}^{-1}[\phi(x,t)]$ where $\hat{\phi}(0) =
\mathcal{F}[\mathcal{CH}[u(x,0)]]$.  In general, for nonlinear PDEs with
initial conditions within a class of functions $\mathcal{U}$, we aim to learn
models that provide predictions $u(t + \tau) = \mathcal{S}_\tau u(t)$
approximating the evolution operator $\mathcal{S}_\tau$ over time-scale $\tau$.
For the Burgers equation, the $\mathcal{CH}$ provides in principle 
an analytic way to obtain a reduced order model by truncating the 
Fourier expansion to $|k| \leq n_f/2$.
This provides for the Burgers equation a benchmark model against which to
compare our learned models. For general PDEs comparable analytic
representations are not usually available.  In addition, for many problems
arising in practice, we are interested primarily in how the system 
behaves for a limited class of initial conditions and parameters. 
The aim then becomes to find a reduced model that predicts well 
the outcomes for these circumstances.  We show how data-driven 
approaches can be developed for this purpose.

We develop VAE methods for learning reduced order models for the responses of
nonlinear Burgers Equation when the initial conditions are from a collection of
functions $\mathcal{U}$.  We consider in particular 
$\mathcal{U} = \mathcal{U}_1 = \{ u\, | \, u(x,t;\alpha) = \alpha 
\sin(2\pi x) + (1-\alpha) \cos^3(2 \pi x)\}$.  We remark that while the initial
conditions are a linear combination of two functions, this relation does 
not hold for the solutions given the non-linear dynamics of the Burgers' 
PDE.  We learn VAE models that extract from $u(x,t)$ latent
variables $z(t)$ to predict $u(x,t + \tau)$.  Given the non-uniqueness of
representations and to promote interpretability of the model, we introduce the
inductive bias that the evolution dynamics in latent space for $z$ is linear of
the form $\dot{z} = -\lambda_0 z$, giving exponential decay rate $\lambda_0$.
For discrete times, we take $z_{n+1} = f_{\theta_\ell}(z_n) = \exp(-\lambda_0
\tau) \cdot z_n$, where $\theta_\ell = (\lambda_0)$.  We treated $\lambda_0$
as a hyperparameter in our studies with $\exp(-\lambda_0 \tau)=0.75$, but this
could also be learned if additional criteria is introduced for the latent-space
representations.  The exponential decay is used to influence how the dynamical 
information is represented in latent space and also helps ensure dynamical 
stability for predictions.  For the encoders and decoders, we consider general
nonlinear mappings which are represented by deep neural networks. 

\begin{figure}[ht]
\centerline{\includegraphics[width=0.78\columnwidth]{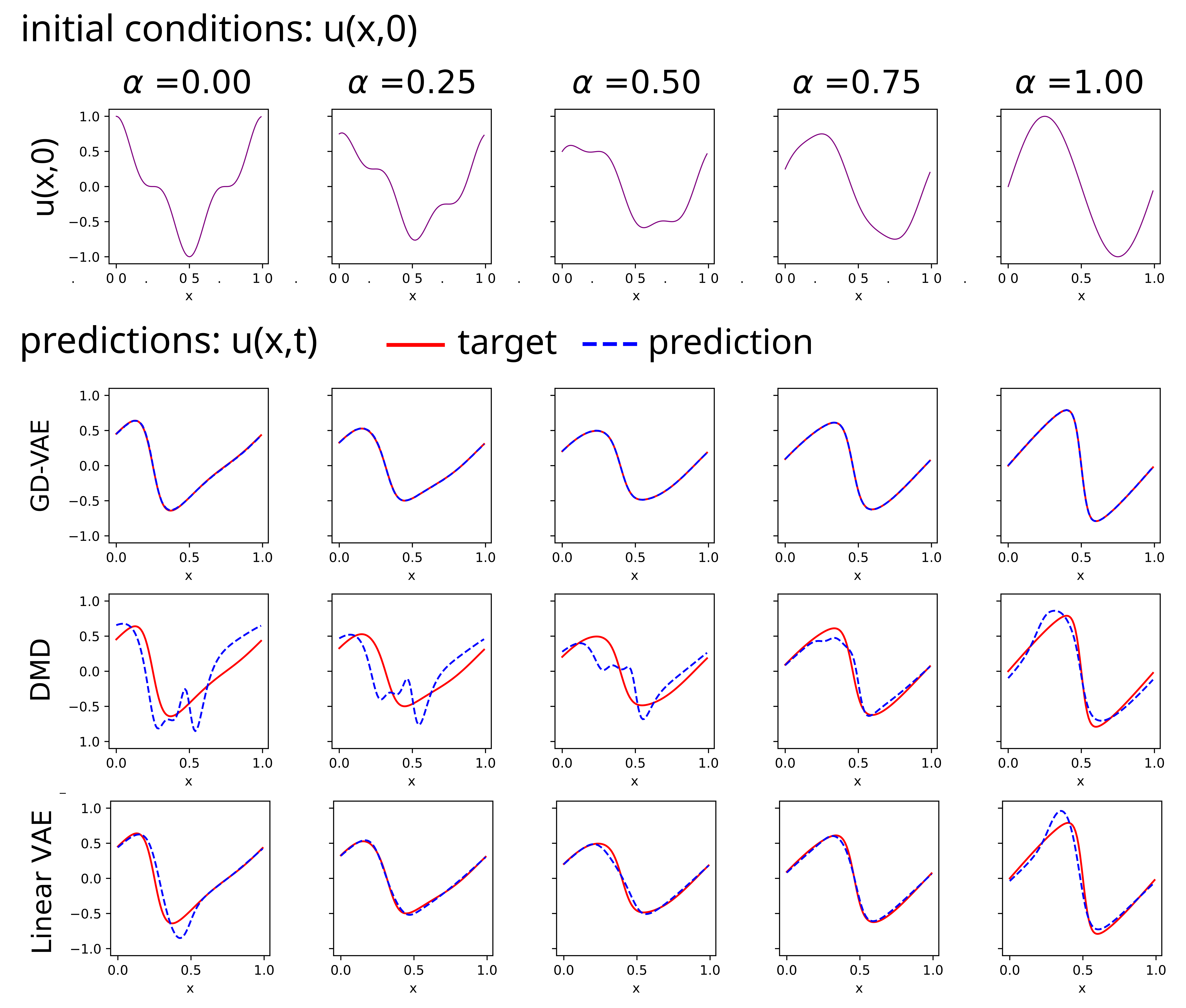}}  
\caption{\textbf{Burgers' Equation: Prediction of Dynamics.}  We consider
responses for initial conditions $\mathcal{U}_1 = \{ u\, | \, u(x,t;\alpha) =
\alpha \sin(2\pi x) + (1-\alpha) \cos^3(2 \pi x)\}$.  Predictions are made for
the evolution $u$ over the time-scale $\tau$ satisfying
equation~\ref{eqn:burgers} with initial conditions in $\mathcal{U}_1$.  We find
our nonlinear VAE methods are able to learn with $2$ latent dimensions the
dynamics with errors $< 1\%$. Methods such as
DMD~\citep{DMD_Schmid_2010,DMD_Theory_and_App_Kutz_2014} with $3$ modes which
are only able to use a single linear space to approximate the initial
conditions and prediction encounter challenges in approximating the nonlinear
evolution.  We find our linear VAE method with $2$ modes provides some
improvements by allowing for using different linear spaces for representing
the input and output functions.
Results are summarized in Table~\ref{table:Burgers_compare_VAE_DMD_etc}.}
\label{fig:Burgers_compare_VAE_DMD_etc}
\end{figure}

We train the model by drawing $m$ samples of 
the prediction pairs $(u^{k_i}(x,t_i),u^{k_i}(x,t_i + \tau))$.
These are sampled from solution 
trajectories with $u^k(x,t) \in \mathcal{S}_{t_i} \mathcal{U}$. The 
$\mathcal{U}$ gives the ensemble of initial conditions. 
The $\mathcal{S}_t$ denotes the non-linear solution operator for the Burgers equation
with $u(x,t) = \mathcal{S}_{t}[u(x,0)]$.  For the $i^{th}$ sample, the  
$k_i$ gives the trajectory index $k$ and the $t_i$ the 
sampled initial time $t$ from the trajectory.  We perform studies 
with parameters $\nu = 2 \times 10^{-2}$, $\tau
= 2.5 \times 10^{-1},m=100$ with our Deep Neural Networks (DNNs) with layer sizes
(in)-400-400-(out), ReLU activations, and $\gamma = 0.5$, $\beta = 1$. The
covariances $\bsy{\Sigma}_e$ and $\bsy{\Sigma}_d$ used are diagonal with
initial standard deviations $\sigma_d = \sigma_e = 4 \times 10^{-3}$ so that
$\bsy{\Sigma}_e = \mbox{diag}(\sigma_e^2)$ and $\bsy{\Sigma}_d =
\mbox{diag}(\sigma_d^2)$.  We show results of our model predictions in
Figure~\ref{fig:Burgers_compare_VAE_DMD_etc} and
Table~\ref{table:Burgers_compare_VAE_DMD_etc}.

\begin{figure}[ht]
\centerline{\includegraphics[width=0.72\columnwidth]{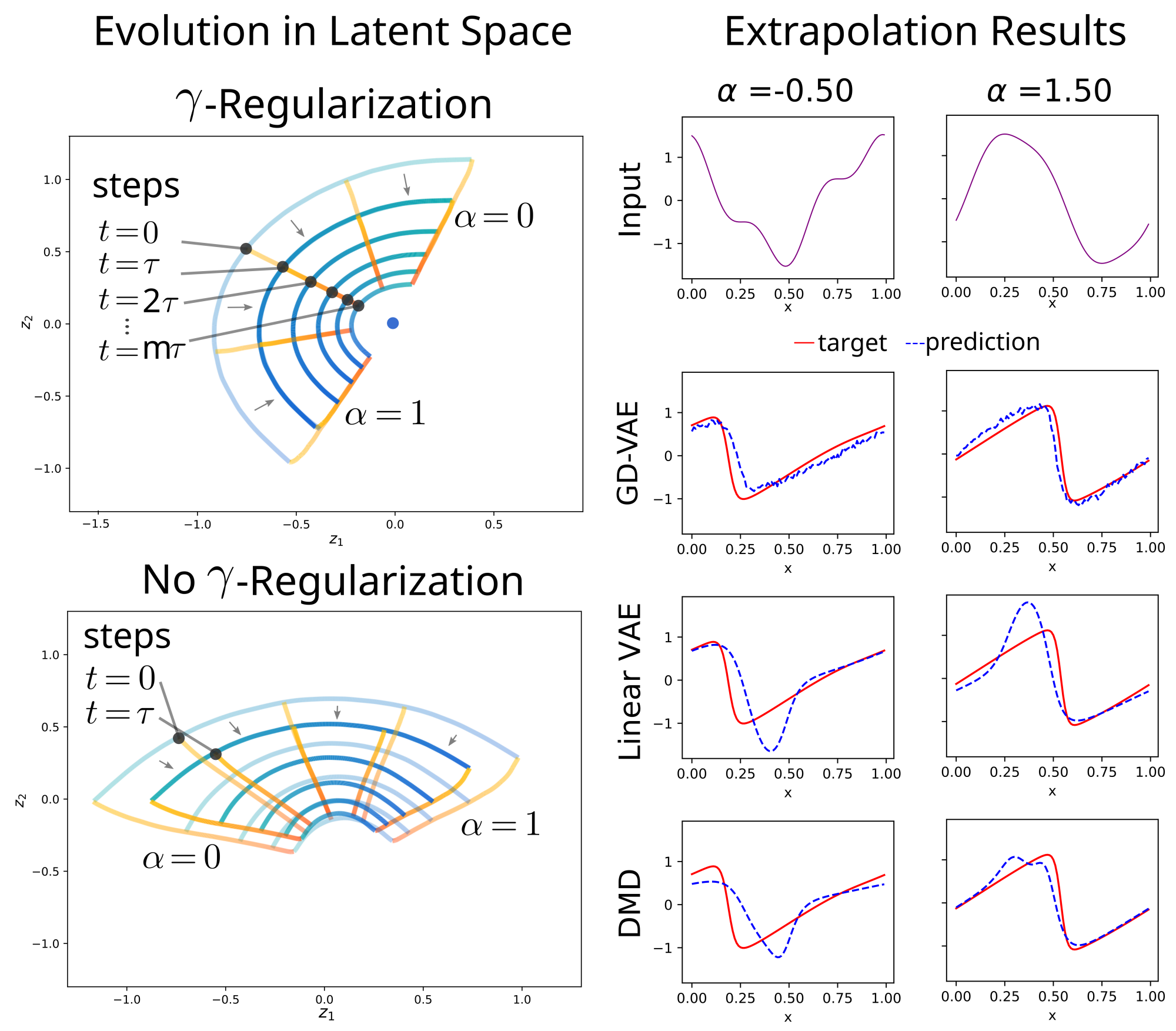}} 
\caption{\textbf{Burgers' Equation: Latent Space Representations and
Extrapolation Predictions.}  We show the latent space representation $z$ of the
dynamics for the input functions $u(\cdot,t;\alpha) \in \mathcal{U}_1$.  VAE
organizes for $u$ the learned representations $z(\alpha,t)$  in parameter
$\alpha$ \textit{(blue-green)} into circular arcs that are concentric in the
time parameter $t$, \textit{(yellow-orange)} \textit{(left)}.  The
reconstruction regularization with $\gamma$ aligns subsequent time-steps of the
dynamics in latent space facilitating multi-step predictions.  The learned VAE
model exhibits a level of extrapolation to predict dynamics even for some
inputs $u \not \in \mathcal{U}_1$ beyond the training data \textit{(right)}.
}
\label{fig:burgers_latent_space}	
\end{figure}

\begin{table}[ht]
\begin{center}
{
\fontsize{9.3}{12}\selectfont
\selectfont
\begin{tabular}{l|c|llll}
\hline 
\rowcolor{atz_table1} 
\textbf{Method} & \textbf{Dim} & \textbf{0.25s} & \textbf{0.50s} & \textbf{0.75s} & \textbf{1.00s} \\
\cline{1-6}
GD-VAE & 2 & $4.27e\sminus 03 \pm 4.8e\sminus 04$ & $4.49e\sminus 03 \pm 4.9e\sminus 04$ & $4.76e\sminus 03 \pm 5.0e\sminus 04$ & $6.43e\sminus 03 \pm 8.1e\sminus 04$ \\
\hline
VAE 2D-Linear & 2 & $1.26e\sminus 01 \pm 2.7e\sminus 04$ & $1.31e\sminus 01 \pm 1.6e\sminus 04$ & $1.24e\sminus 01 \pm 2.1e\sminus 04$ & $1.26e\sminus 01 \pm 2.4e\sminus 04$ \\
DMD-3D & 3 & $2.21e\sminus 01\pm 0.0e\sminus 00 $ & $1.79e\sminus 01\pm 0.0e\sminus 00 $ & $1.56e\sminus 01\pm 0.0e\sminus 00 $ & $1.49e\sminus 01\pm 0.0e\sminus 00 $ \\
POD-3D & 3 & $3.24e\sminus 01\pm 0.0e\sminus 00 $ & $4.28e\sminus 01\pm 0.0e\sminus 00 $ & $4.87e\sminus 01\pm 0.0e\sminus 00 $ & $5.41e\sminus 01\pm 0.0e\sminus 00 $ \\
\hline
Cole\sminus Hopf-2D & 2 & $5.18e\sminus 01\pm 0.0e\sminus 00 $ & $4.17e\sminus 01\pm 0.0e\sminus 00 $ & $3.40e\sminus 01\pm 0.0e\sminus 00 $ & $1.33e\sminus 01\pm 0.0e\sminus 00 $ \\
Cole\sminus Hopf-4D & 4 & $5.78e\sminus 01\pm 0.0e\sminus 00 $ & $6.33e\sminus 02\pm 0.0e\sminus 00 $ & $9.14e\sminus 03\pm 0.0e\sminus 00 $ & $1.58e\sminus 03\pm 0.0e\sminus 00 $ \\
Cole\sminus Hopf-6D & 6 & $1.48e\sminus 01\pm 0.0e\sminus 00 $ & $2.55e\sminus 03\pm 0.0e\sminus 00 $ & $9.25e\sminus 05\pm 0.0e\sminus 00 $ & $7.47e\sminus 06\pm 0.0e\sminus 00 $ \\
\hline
\end{tabular} \\
}
\end{center}
\caption{\textbf{Burgers' Equation: Prediction Accuracy.} 
The reconstruction $L^1$-relative errors in predicting $u(x,t)$ and 
standard error over 5 training trials.  We compare our VAE
methods, Dynamic Model Decomposition (DMD), and Proper Orthogonal
Decomposition (POD), and reduction by Cole-Hopf (CH), over multiple-steps and
number of latent dimensions (Dim).
}
\label{table:Burgers_compare_VAE_DMD_etc}
\end{table}

We investigate the importance of the non-linear approximation properties of our 
methods in capturing system behaviors.  We do this by making comparisons with linear
methods that include Dynamic Mode
Decomposition (DMD)~\citep{DMD_Schmid_2010,DMD_Theory_and_App_Kutz_2014},
Proper Orthogonal Decomposition (POD)~\citep{POD_Intro_Chatterjee_2000}, and
a linear variant of our GD-VAE approach.  Recent CNN-AEs have also studied related
advantages of non-linear
approximations~\citep{Carlberg_Lee_Nonlinear_Dynamics_AE_2020}.  Some
distinctions in our work is the use of VAEs to further regularize the autoencoders and 
using topological latent spaces to facilitate further capturing inherent structures.  The
DMD and POD are widely used and successful approaches that aim to find an
optimal linear space on which to project the dynamics and learn a linear
evolution law for system behaviors.  DMD and POD have been successful in
obtaining models for many applications, including steady-state fluid mechanics
and transport problems \citep{DMD_Theory_and_App_Kutz_2014, DMD_Schmid_2010}.
However, given their inherent linear approximations they can encounter
well-known challenges related to translational and rotational invariances, as
arise in advective phenomena and other
settings~\citep{Kutz_Brunton_book_ch_ROMs_2019}.    Our comparison studies can
be found in Table~\ref{table:Burgers_compare_VAE_DMD_etc}.

\begin{table}[h]
\begin{center}
{
\fontsize{9.3}{12}\selectfont
\begin{tabular}{c|ccccc}
\hline
\rowcolor{atz_table1} 
$\bsy{\gamma}$ & \textbf{0.00s} & \textbf{0.25s} & \textbf{0.50s} & \textbf{0.75s} & \textbf{1.00s} \\
\cline{1-6}
0.00 & $1.20e\sminus 01 \pm 5.8e\sminus 03$ & $3.77e\sminus 03 \pm 6.9e\sminus 04$ & $9.38e\sminus 02 \pm 6.7e\sminus 03$ & $1.84e\sminus 01 \pm 1.3e\sminus 02$ & $2.83e\sminus 01 \pm 1.9e\sminus 02$ \\
0.05 & $4.18e\sminus 03 \pm 4.0e\sminus 04$ & $3.80e\sminus 03 \pm 4.0e\sminus 04$ & $4.00e\sminus 03 \pm 4.0e\sminus 04$ & $4.24e\sminus 03 \pm 4.1e\sminus 04$ & $4.71e\sminus 03 \pm 4.7e\sminus 04$ \\
2.00 & $5.20e\sminus 03 \pm 4.8e\sminus 04$ & $5.03e\sminus 03 \pm 5.1e\sminus 04$ & $5.30e\sminus 03 \pm 4.8e\sminus 04$ & $5.55e\sminus 03 \pm 4.9e\sminus 04$ & $8.52e\sminus 03 \pm 1.1e\sminus 03$ \\
\hline
\end{tabular}
} 
\end{center}
\caption{\textbf{Burgers' Equation: Prediction Accuracy and Regularizations.} 
The reconstruction $L^1$-relative errors and standard error over 5 training
trials. We show prediction of $u(x,t)$ varying the strength of the
reconstruction regularization $\gamma$ in equation~\ref{equ:vae_loss}.
}
\label{table:Burgers_gamma_beta}
\end{table}

We also considered how our VAE methods performed when adjusting the parameter
$\gamma$ for the strength of the
reconstruction regularization.  The reconstruction regularization has a
significant influence on how the VAE organizes representations in latent space
and the accuracy of predictions of the dynamics, especially over multiple
steps, see Figure~\ref{fig:burgers_latent_space} and
Table~\ref{table:Burgers_gamma_beta}.     The regularization serves to
align representations consistently in latent space facilitating multi-step
compositions.  We also found our VAE learned representations capable of some
level of extrapolation beyond the training data, see 
Table~\ref{table:Burgers_gamma_beta}.  While extrapolation was not our
main aim, it is interesting that the learned neural network representations
appear to be based on features that retain enough information to be able to predict
system responses reasonably even outside the training domain.

\subsection{Burgers' Equation: Topological Structure of Parameterized PDEs}
For parameterized PDEs and other systems, there is often significant prior
geometric and topological information available concerning the space
of responses of the system.  For example, the parameters may vary periodically
or have other imposed constraints limiting possible responses.  We show 
how our methods can be utilized to make use of this structure to 
learn latent space representations having specified topological 
or geometric structures.  

\subsection{Periodic and Doubly-Periodic Systems: Cylinder and Torus Latent Spaces}
\label{sec_burgers_topo}

\begin{figure}[h]
\centerline{\includegraphics[width=0.9\columnwidth]{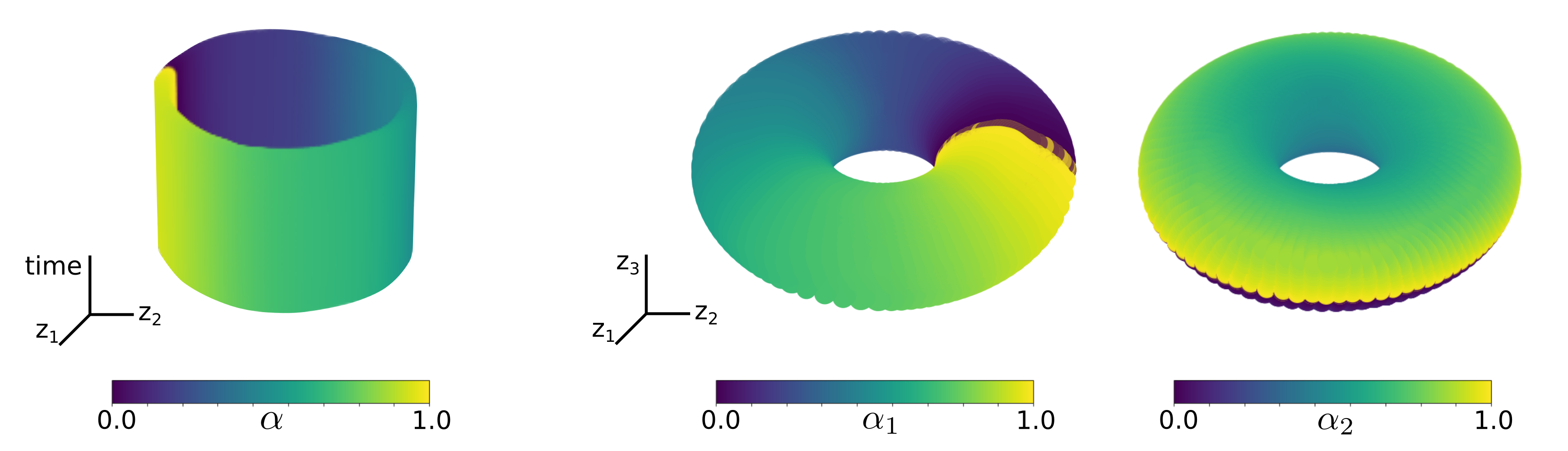}}
\caption{\textbf{GD-VAE Manifold Latent Spaces for Periodic and Doubly-Periodic
Systems.} 
We show results for
the learned mappings of solutions of the parameterized 
Burgers Equation in the periodic $\alpha$ case and doubly-periodic 
$\alpha_1,\alpha_2$ case. 
For such time-dependent PDE solutions, it is natural to consider latent
spaces with abstract cylinder-like shapes represented as a 
product-space $\mathcal{B} \times \mathbb{R}$.  In this cylinder topology, 
the infinite axis $\mathbb{R}$ corresponds to time and the 
other dimensions to the state of the system for 
some base space $\mathcal{B}$.  For a single periodic parameter, the
base-space $\mathcal{B}$ is a circle giving the conventional 
cylinder shown on the \textit{(left)}.
For two doubly-periodic parameters, it is natural to consider
a torus for the base-space $\mathcal{B}$.  In practice, we 
find in training there are benefits to using the 
Clifford Torus $S^1 \times S^1 \subset \mathbb{R}^4$.  For visualization
purposes, we depict our results mapping this onto the conventional torus in 
$\mathbb{R}^3$ as shown on the \textit{(right)}.  }
\label{fig_periodic_double_periodic}
\end{figure}

We start by considering Burgers' equations with initial conditions 
parameterized periodically as 
\begin{equation}
\label{equ_burgers_periodic_init}
u_{\alpha}(x,t=0) = \begin{bmatrix} \cos(2 \pi \alpha) \\ 
\sin(2 \pi \alpha) \end{bmatrix} 
\cdot 
\begin{bmatrix} \cos(2 \pi x) \\ \sin(2 \pi{x}) 
\end{bmatrix}, 
  \quad\quad \alpha \in [0,1].
\end{equation}
We consider solutions with $x,t \in [0,1]$.  Since the boundary conditions
are periodic $u(0,t) = u(1,t)$, the initial conditions parameterized by $\alpha$
effectively shift relative to one another and we have the topology of a circle.  
We also consider initial conditions parameterized doubly-periodic as
\begin{equation}
\label{equ_burgers_doubly_periodic_init}
u_{\alpha_1, \alpha_2}(x, 0) = \begin{bmatrix} 
\cos(\alpha_1) \\ 
\sin(\alpha_1) \\ 
\cos(\alpha_2) \\ 
\sin(\alpha_2) 
\end{bmatrix} \cdot 
\begin{bmatrix} 
\cos(2 \pi x) \\ 
\sin(2 \pi x) \\ 
\cos(4 \pi x) \\ 
\sin(4 \pi x) 
\end{bmatrix}, 
\quad \alpha_1, 
\alpha_2 \in [0,2\pi].
\end{equation}
This corresponds to the topology of a torus.  We can project our solutions onto
the Clifford Torus given by the product space $S^1 \times S^1 \subset
\mathbb{R}^4$, where $S^1$ is the circle space~\citep{Killing1885,Volkert2013}.
For example representations of latent spaces, see
Figure~\ref{fig_periodic_double_periodic}.  In each case, we seek
representations that disentangle state information of
the collection of solutions $\{u_\alpha\}_{0 \leq \alpha \leq 1}$ 
or $\{u_{\alpha_1,\alpha_2}\}_{0 \leq \alpha_1,\alpha_2 \leq 2\pi}$.

In principle, for the periodic case, one analytic way the family of solutions
$\{u_{{\alpha}}\}_{0 \leq \alpha \leq 1}$ could be encoded continuously is to use 
the $\mathbb{R}^2$ latent space with mapping
\begin{equation}
\label{equ_rep_cone_proj}
u_{\alpha}(x, t) \rightarrow \left(t \cos(2\pi \alpha), t \sin(2\pi \alpha)
\right).
\end{equation}
This is similar to projecting a cone-shaped manifold in three dimensions to the
two dimensional $xy$-plane.  However, in practice, this is hard to learn from
the training data given how this representation entangles the time 
and configuration state information, 
see Appendix~\ref{sec_role_geo_struc}. 
Also, while such encodings may be useful for some problems, using 
models that disentangle time and the configuration state information 
into separate dimensions in the latent space is expected to result in 
simpler manifold structures and more interpretable results.  
We impose this structure by seeking representations of the form
\begin{equation}
\nonumber
u_{\bsy{\alpha}}(x, t) \rightarrow \left(\mb{f}(\bsy{\alpha}),t\right).
\end{equation}
The mapping $\mb{f}$ does not depend on time and captures the current state information. 
As a consequence of the potentially non-trivial topology of the $\bsy{\alpha}$ latent 
variable, Euclidean latent spaces with a continuous encoder 
that disentangles time must also map to a sufficiently larger number 
of dimensions to embed the manifold to preserve the topology.
In the periodic case, this would be to at least three 
dimensions.  An example of such an analytic mapping to $\mathbb{R}^3$ is
\begin{equation}
\nonumber
u_{\alpha}(x, t) \rightarrow \left(\cos(2\pi \alpha), \sin(2\pi \alpha),t\right).
\end{equation}
This mapping corresponds in a latent space having a cylinder topology.
Similarly, for the doubly-periodic family $\{u_{\alpha_1,\alpha_2}\}$ the state
information can be mapped in principle to a torus by $\mb{f}(\bsy{\alpha})$.
This would yield the disentangled embedding $u_{\bsy{\alpha}}(x, t)
\rightarrow \left(\mb{f}(\bsy{\alpha}),t\right)$ with 
$\bsy{\alpha} = (\alpha_1,\alpha_2)$.  The torus can be represented in principle
in $\mathbb{R}^3$ or as a product space $S^1\times
S^1$ where $S^1$ is the circle space.  This latter representation
is referred to as the Clifford Torus~\citep{Killing1885,Volkert2013}.  This
gives the latent space with the cylinder topology $\mathcal{B}\times\mathbb{R}$,
where $\mathcal{B}$ is the base-space.  For periodic case the $\mathcal{B} = S^1$ is
a circle and for the doubly-periodic case the $\mathcal{B} = S^1\times S^1$ 
is a torus. 

We demonstrate how our methods can be used in practice to learn representations
for the dynamics of the Burgers' Equation solutions for latent spaces having
specified topologies.  We use a few approaches to leverage the prior geometric
information about the system and to regulate how the model organizes the latent
representations.  This includes (i) specifying an explicit time-evolution map
to disentangle time by requiring $f_{ev}(\mb{z}) = \mb{z} + \Delta{t}\mb{e}_n
\in \mathbb{R}^n$, where $\Delta{t}$ is the discrete time step and $\mb{e}_n
=(0,\ldots,0,1)$ is the $n^{th}$-standard basis vector.  We also use (ii)
projection of the representation to a manifold structure, here we project the
first $n-1$ dimensions to impose the topology as discussed in 
Section~\ref{sec_manifold_latent_spaces}.
In this way, we can introduce inductive biases for general initial conditions
leveraging topologic and geometric knowledge to regulate the latent
representations toward having desirable properties. 
 
For our example of the Burger's Equations~\ref{equ_burgers_equ} with the
periodic initial conditions in equation~\ref{equ_burgers_periodic_init}, we train using
$10^4$ solutions sampled uniformly over $\alpha$ and time $t$.  The samples are
corrupted with a Gaussian noise having standard deviation $\sigma=0.02$.  We
train GD-VAEs using our framework discussed in
Section~\ref{sec_gd_vae}.  The neural networks have the architecture with
number of layers for encoders {(in)}-$400$-$400$-{(latent-space)}
and decoders {(latent-space)}-$400$-$400$-{(out)}.  The layers
have a bias except for the last layer and {ReLU} activation
functions~\citep{Goodfellow2016,LeCun2015}.  The numerical solutions of the
Burgers equation~\ref{equ_burgers_equ} are sampled as $u(x_k,t) = u_k$ with
$x_k$ at $n=100$ points giving {(in)}={(out)}=$n$=$100$. We also perform 
trainings representing the encoder variance with a
trainable deep neural network, and set decoder variance such that log
likelihood can be viewed as a Mean Square Error (MSE) loss with weight 
$10^{-4}$.

We predict the future evolution of the solutions of
the Burger's Equation over multiple steps using equation~\ref{sec_gd_vae}.  
We investigate the $L^1$-accuracy of the learned GD-VAE predictions relative 
to the numerical solutions.  We compare the GD-VAE with more conventional VAE 
methods that do not utilize topological information.  We also make
comparisons with standard Autoencoders (AEs) with 
and without utilizing topological information and our projection approaches 
in Section~\ref{sec_manifold_latent_spaces}.  We refer in the notation 
to our geometric projection as \textit{g-projection}.  
We show our results in Table~\ref{table_burgers_topo_predict}.

We find our GD-VAE methods are able to learn parsimoneous disentangled
representations that achieve a comparable reconstruction accuracy to standard
VAE and AE provided these latter methods use enough dimensions.  Reconstruction
accuracy alone is only one way to measure the quality of the latent-space
representations.  Topological considerations play a significant role.  When
using too few dimensions, we see the standard AE and VAE methods can struggle
to learn a suitable representation, see Figure~\ref{fig_burgers_topo_space}.
As we discuss in more detail in Appendix~\ref{sec_role_geo_struc}, this arises
since the autoencoders involve continuous maps that are unable to accomodate 
the topology injectively.  For periodic systems this can result in rapid 
looping back behaviors approximating a discontinuity which results in 
scattered latent space points, see Figure~\ref{fig_topo_incompatible}.
For additional discussions of the role of latent topology and 
training behaviors see Appendix~\ref{sec_role_geo_struc}.

For longer-time multi-step predictions, we find our geometric projections can
help with stability of the predictions arising from composition of the learned
maps.  We find both GD-VAEs and AEs with \textit{g-projection} have enhanced
stability when compared to standard AEs for multi-step predictions.  We find at
time $t=1.00s$, the AE+\textit{g-projection} still is able to make accurate
predictions, while the standard AE incurs significant errors, see
Table~\ref{table_burgers_topo_predict}.  This arises from the geometric
constraints reducing the number of dimensions and the "volume" of 
the subset of points in the latent space overwhich the encoder and decoders must 
learn correct mappings for responses.  The geometric projection 
also serves to enhance the
statistical power of the training data given more opportunities for common
overlap of cases.  In contrast, for multi-step predictions by general autoencoder maps
new latent-space codes can arise during compositions 
that are far from those encountered in the training set
resulting in unknown behaviors for the dynamic predictions. 
While in some cases there is comparable $L^1$-reconstruction 
errors to AEs with \textit{g-projection}, the VAEs give
better overall representations and more reliable training 
since the noise regularizations 
result in smoother organization of the latent codes with less
local sensitivity in the encoder-decoder maps relative to 
the AEs.  The geometric projections developed for GD-VAEs provide further 
benefits for stability by constraining the latent-space dynamics to be 
confined within a closed manifold which further
enhances the multi-step predictions, see results for $t=1.00s$ in 
Table~\ref{table_burgers_topo_predict}. 

\begin{table}[!h]
\begin{center}
{\fontsize{6.3}{8}\selectfont
\selectfont
\begin{tabular}{l|r|lllll}
\hline 
\rowcolor{atz_table2} 
  \multicolumn{7}{l}{\textbf{Periodic: Cylinder Topology.}}\\
\hline
\rowcolor{atz_table1} 
\textbf{Method} & \textbf{Dim} & \textbf{0.00s} & \textbf{0.25s} & \textbf{0.50s} & \textbf{0.75s} & \textbf{1.00s}\\
\hline 
GD-VAE & 3 & $2.12e\sminus 02 \pm 9.3e\sminus 05$ & $2.15e\sminus 02 \pm 1.5e\sminus 03$ & $2.57e\sminus 02 \pm 2.7e\sminus 03$ & $3.14e\sminus 02 \pm 4.0e\sminus 03$ & $4.72e\sminus 02 \pm 5.5e\sminus 03$ \\
\hline
VAE-2D & 2 & $2.51e\sminus 02 \pm 1.6e\sminus 03$ & $2.33e\sminus 02 \pm 1.9e\sminus 03$ & $2.93e\sminus 02 \pm 2.3e\sminus 03$ & $4.01e\sminus 02 \pm 2.9e\sminus 03$ & $6.42e\sminus 02 \pm 4.5e\sminus 03$ \\
VAE-3D & 3 & $2.32e\sminus 02 \pm 3.8e\sminus 03$ & $2.38e\sminus 02 \pm 2.9e\sminus 03$ & $2.98e\sminus 02 \pm 3.4e\sminus 03$ & $3.78e\sminus 02 \pm 4.0e\sminus 03$ & $5.67e\sminus 02 \pm 5.6e\sminus 03$ \\
VAE-10D & 10 & $1.99e\sminus 02 \pm 5.9e\sminus 04$ & $1.99e\sminus 02 \pm 8.8e\sminus 04$ & $2.49e\sminus 02 \pm 1.2e\sminus 03$ & $3.20e\sminus 02 \pm 1.6e\sminus 03$ & $4.88e\sminus 02 \pm 2.8e\sminus 03$ \\
\hline
AE (g-projection) & 3 & $1.45e\sminus 02 \pm 7.7e\sminus 04$ & $1.47e\sminus 02 \pm 8.1e\sminus 04$ & $1.57e\sminus 02 \pm 9.0e\sminus 04$ & $1.70e\sminus 02 \pm 9.8e\sminus 04$ & $1.98e\sminus 02 \pm 9.7e\sminus 04$ \\
AE (no projection) & 3 & $1.39e\sminus 02 \pm 4.6e\sminus 04$ & $1.39e\sminus 02 \pm 6.0e\sminus 04$ & $8.55e\sminus 02 \pm 5.9e\sminus 03$ & $1.99e\sminus 01 \pm 1.1e\sminus 02$ & $3.18e\sminus 01 \pm 1.3e\sminus 02$ \\
\hline
\end{tabular} \\
}
\vspace{0.4cm}
{\fontsize{6.3}{8}\selectfont
\selectfont
\begin{tabular}{l|r|lllll}
\hline 
\rowcolor{atz_table2} 
  \multicolumn{7}{l}{\textbf{Doubly-Periodic: Torus Topology.}}\\
\hline
\rowcolor{atz_table1} 
\textbf{Method} & \textbf{Dim} & \textbf{0.00s} & \textbf{0.25s} & \textbf{0.50s} & \textbf{0.75s} & \textbf{1.00s}\\
\hline 
GD-VAE & 5 & $5.48e\sminus 02 \pm 3.8e\sminus 03$ & $4.75e\sminus 02 \pm 4.4e\sminus 03$ & $7.32e\sminus 02 \pm 5.8e\sminus 03$ & $1.06e\sminus 01 \pm 7.3e\sminus 03$ & $1.55e\sminus 01 \pm 1.0e\sminus 02$ \\ 
\hline
VAE-3D & 3 & $2.37e\sminus 01 \pm 2.9e\sminus 03$ & $2.22e\sminus 01 \pm 3.3e\sminus 03$ & $4.09e\sminus 01 \pm 1.5e\sminus 02$ & $6.39e\sminus 01 \pm 2.1e\sminus 02$ & $7.97e\sminus 01 \pm 2.6e\sminus 02$ \\ 
VAE-5D & 5 & $5.56e\sminus 02 \pm 1.3e\sminus 03$ & $4.52e\sminus 02 \pm 1.8e\sminus 03$ & $6.73e\sminus 02 \pm 2.7e\sminus 03$ & $9.66e\sminus 02 \pm 4.0e\sminus 03$ & $1.42e\sminus 01 \pm 5.7e\sminus 03$ \\ 
VAE-10D & 10 & $5.45e\sminus 02 \pm 2.1e\sminus 03$ & $4.61e\sminus 02 \pm 2.0e\sminus 03$ & $7.12e\sminus 02 \pm 2.2e\sminus 03$ & $1.01e\sminus 01 \pm 2.6e\sminus 03$ & $1.47e\sminus 01 \pm 4.6e\sminus 03$ \\ 
\hline  
AE (g-projection) & 5 & $5.42e\sminus 02 \pm 6.4e\sminus 03$ & $5.02e\sminus 02 \pm 5.9e\sminus 03$ & $6.35e\sminus 02 \pm 9.2e\sminus 03$ & $8.12e\sminus 02 \pm 1.3e\sminus 02$ & $1.03e\sminus 01 \pm 1.8e\sminus 02$ \\ 
AE (no projection) & 5 & $2.80e\sminus 02 \pm 5.4e\sminus 04$ & $3.16e\sminus 02 \pm 3.9e\sminus 04$ & $1.84e\sminus 01 \pm 6.9e\sminus 03$ & $4.59e\sminus 01 \pm 3.1e\sminus 02$ & $8.04e\sminus 01 \pm 6.3e\sminus 02$ \\ 
\hline
\end{tabular}
}
\end{center}
\caption{\textbf{Burgers' Equations for Periodic and Doubly-Periodic
Parameterized PDEs: $L^1$-Prediction Accuracy.} The $L^1$-relative errors 
and standard error over 5 training trials.  We train using the  GD-VAEs
framework discussed in Sections~\ref{sec_gd_vae}
and~\ref{sec_manifold_latent_spaces}.  We make comparisons to more conventional
Variational Autoencoders (VAEs) and standard Autoencoders (AEs).  The
AE-(g-projection) method refers to the case where we combine using a 
standard AE (non-variational) and project the 
latent codes using our geometric projection approach developed in
Section~\ref{sec_manifold_latent_spaces}.  
}
\label{table_burgers_topo_predict}
\end{table}

\begin{figure}
\centerline{
\includegraphics[width=0.99\columnwidth]{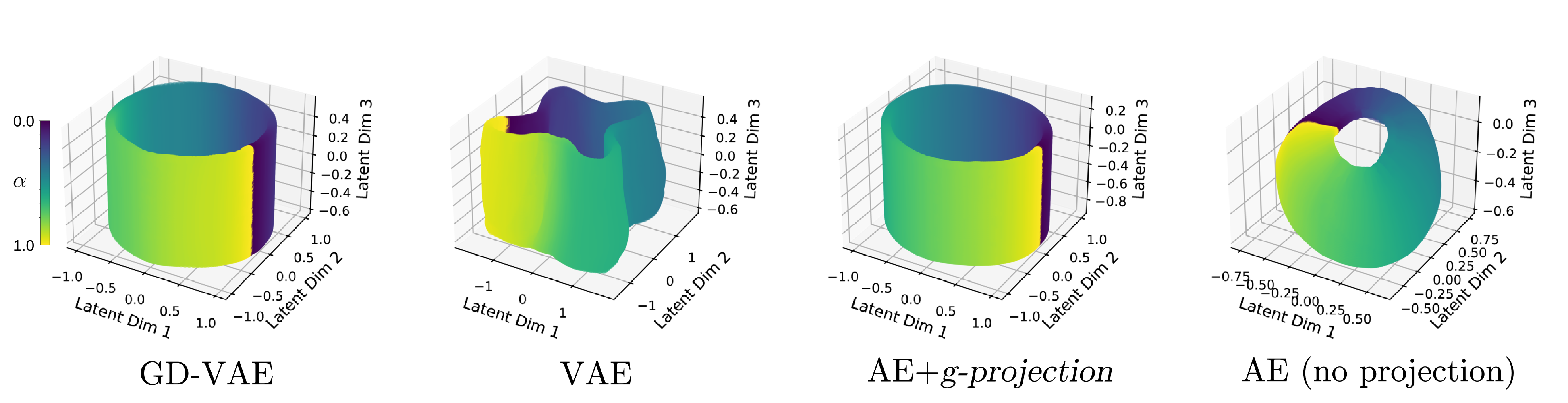}    
}
\caption{\textbf{Learned Latent Spaces for the Burgers' Equations.} 
The methods learn different three dimensional latent space representations for
the non-linear dynamics.  The GD-VAE method uses a model that encodes
state information of the solutions on a circle which is aligned temporally 
\textit{(left)}.  This allows for a representation of the evolution dynamics 
in the latent space following a simple contour in the direction of 
the Latent Dimension $3$.  While VAE without using a
geometric prior learns an embedding having cylindrical topology the latent
representation is geometrically distorted \textit{(left-middle)}.  
The AE+\textit{g-projection} is found to learn a latent space representation 
with disentangled time and state information and with encodings 
that are aligned temporally \textit{(right-middle)}.  The AE without 
using a geometric prior learns an
arbitrary embedding that entangles the temporal and state information
\textit{(right)}.  The general AE representations also do not have a preferred scale
nor an alignment with the coordinate origin relative to the GD-VAE methods.  
The \textit{g-projection} approach is discussed in more detail in
Section~\ref{sec_manifold_latent_spaces}. 
}
\label{fig_burgers_topo_space}
\end{figure}

\newpage
\clearpage
\subsection{Constrained Mechanical Systems: Learning with General Manifold Latent Spaces}

\begin{figure}[ht]
\centerline{\includegraphics[width=0.9\columnwidth]{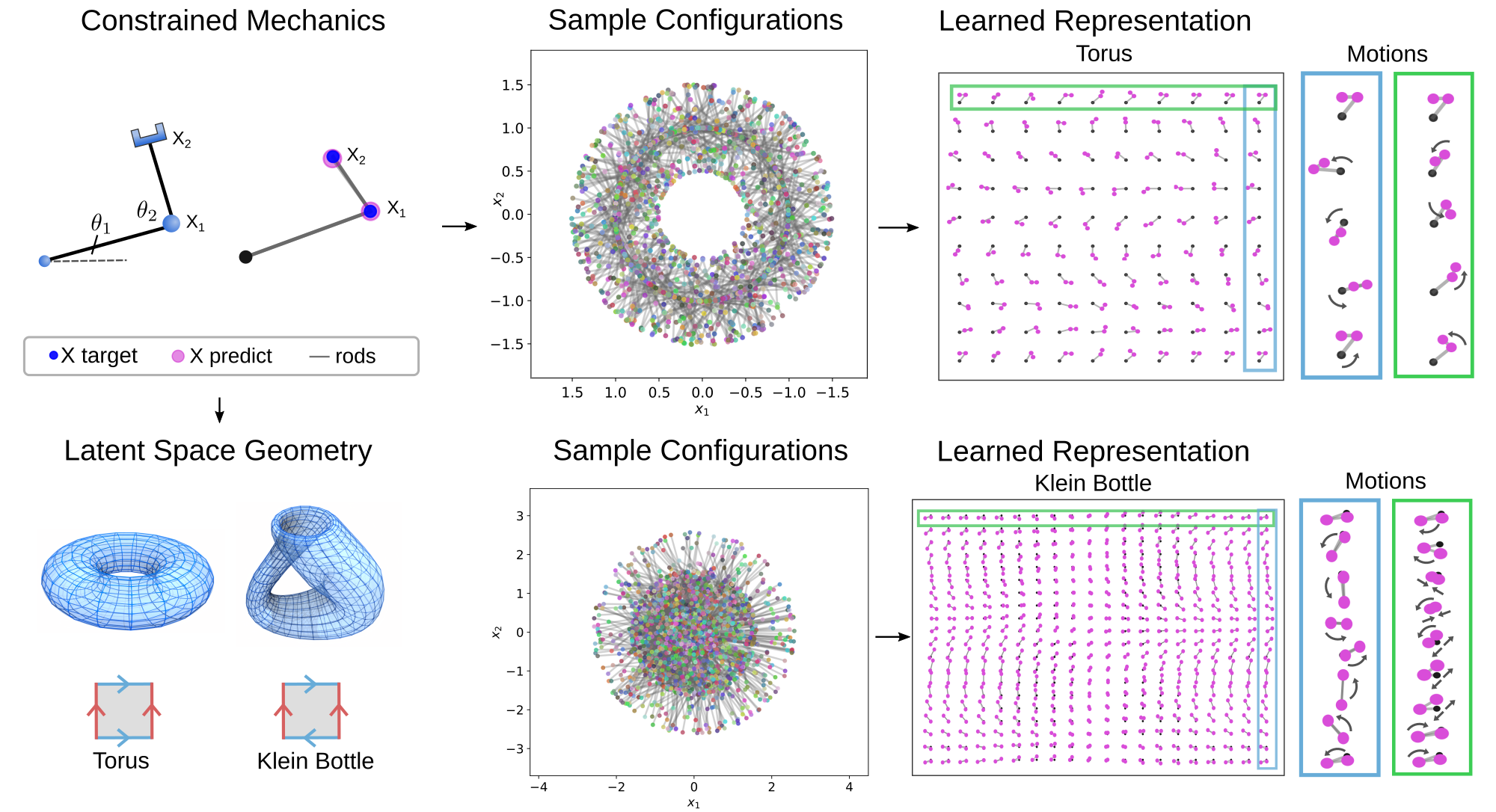}}
\caption{\textbf{GD-VAE Representations of Motions using Manifold Latent Spaces.}
We learn from observations representations for constrained mechanical systems
using manifold latent spaces of general shape $\mathcal{M}$.  The arm
mechanism has configurations $\mb{x} = (\mb{x}_1,\mb{x}_2) \in \mathbb{R}^4$
\textit{(left)}.
For rigid segments, the motions are constrained to be on a manifold (torus)
$\mathcal{M} \subset \mathbb{R}^4$.  For extendable segments, we can also
consider more exotic constraints, such as requiring $\mb{x}_1,\mb{x}_2$ to be
on a klein bottle in $\mathbb{R}^4$ \textit{(middle)}.  Results of our GD-VAE
methods for learned representations for motions under these constraints are
shown on the \textit{(right)}.  GD-VAE learns the segment length constraint 
and two nearly decoupled coordinates for the torus data that mimics 
the roles of angles \textit{(top-right)}.  For the klein bottle case,
GD-VAEs learns two rotation-like motions to generate 
the constrained configurations
\textit{(bottom-right)}.  }
\label{fig:manifold_lvms}
\end{figure}

We consider physical systems with constrained mechanics, such as the 
arm mechanism for reaching for objects in Figure~\ref{fig:manifold_lvms}. 
The observations are taken to be the two
locations $\mb{x}_1, \mb{x}_2 \in \mathbb{R}^2$ giving $\mb{x} =
(\mb{x}_1,\mb{x}_2) \in \mathbb{R}^4$.  When the segments are 
constrained, these configurations lie on a manifold embedded 
in $\mathbb{R}^4$.  The aim of these studies is to show
how GD-VAEs with manifold latent spaces can be used to 
learn representations for constrained systems.  We do not
currently consider dynamical predictions in these studies.
This corresponds to GD-VAEs with latent space dynamics  
$\mb{z}' = f_{\theta_\ell}(\mb{z}) = \mb{z}$ given by the
identity map.
We focus instead on the types of representations learned 
for the constrained latent geometries.   When 
the two segments are constrained to be rigid,
this results in the collection of configurations having the 
topology of a torus.  We can also allow
the segments to extend and consider more exotic constraints.  For example, we could 
require the two points $\mb{x}_1, \mb{x}_2$ always be within a constraint set 
that is a klein bottle surface within $\mathbb{R}^4$.
Related situations arise in other areas of imaging and mechanics, such as in
pose estimation and in studies of visual perception
~\citep{Carlsson_Klein_Bottle_Images_2014, Carlsson2008,
Sarafianos_Review_Pose_Estimation_2016}.  

For the arm mechanics, we can use
this prior knowledge to construct a latent space having the topology of a torus
represented by the product space of two 
circles $S^1 \times S^1$.  To obtain a learnable class of
manifold encoders, we use the family of maps $\mathcal{E}_\theta =
\Lambda(\tilde{\mathcal{E}}_\theta(x))$.  We take $\tilde{\mathcal{E}}_\theta(x)$
to map into $\mathbb{R}^4$ and $\Lambda(\mb{w}) = \Lambda(w_1,w_2,w_3,w_4) =
(z_1,z_2,z_3,z_4) = \mb{z}$, where $(z_1,z_2) = (w_1,w_2)/\|(w_1,w_2)\|$ and
$(z_3,z_4) = (w_3,w_4)/\|(w_3,w_4)\|$.  This provides an analytic formulation we can use
to perform gradient-based learning in combination with our back-propagation training 
methods in
Section~\ref{sec_manifold_latent_spaces}.  For the case of klein bottle constraints, 
we use a numerical approach based on a point-cloud representation of the 
non-orientable manifold with parameterized embedding in $\mathbb{R}^4$ given by
\begin{eqnarray*}
\begin{array}{ll}
z_1 = (a + b\cos(u_2))\cos(u_1) & 
z_2 = (a + b\cos(u_2))\sin(u_1) \\
z_3 = b\sin(u_2)\cos\left(\frac{u_1}{2}\right) &
z_4 = b\sin(u_2)\sin\left(\frac{u_1}{2}\right).
\end{array}
\end{eqnarray*}
We take $u_1,u_2 \in [0,2\pi]$.  The $\Lambda(\mb{w})$ is taken to be the map to
the nearest point of the manifold $\mathcal{M}$, which we use to compute 
numerically the surface information. This allows for gradient-based learning when 
combined with our back-propagation training methods discussed in 
Section~\ref{sec_manifold_latent_spaces}.

We train our GD-VAE methods with encoder and decoder DNNs having layers of
sizes (in)-100-500-100-(out) with Leaky-ReLU activations with $s = 1e\sminus 6$
with results reported in Figure~\ref{fig:manifold_lvms}.  The results
demonstrate how we can learn smooth representations for constrained mechanical
systems for both orientable and non-orientable manifold latent spaces.  For
both the torus and klein bottle latent spaces, we see smooth representations
are learned for generating the configurations.  For the latent torus manifold,
the representation is comparable to the two angular degrees of freedom.  This
shows our GD-VAE approaches can be used as unsupervised learning methods for
contrained mechanical systems to learn representations in manifold latent
spaces with general topology and orientability.

\subsection{Reaction-Diffusion PDEs in 2D: Learning Representations 
for Spatially Distributed Dynamics}

\begin{figure}[ht]
\centerline{\includegraphics[width=0.9\columnwidth]{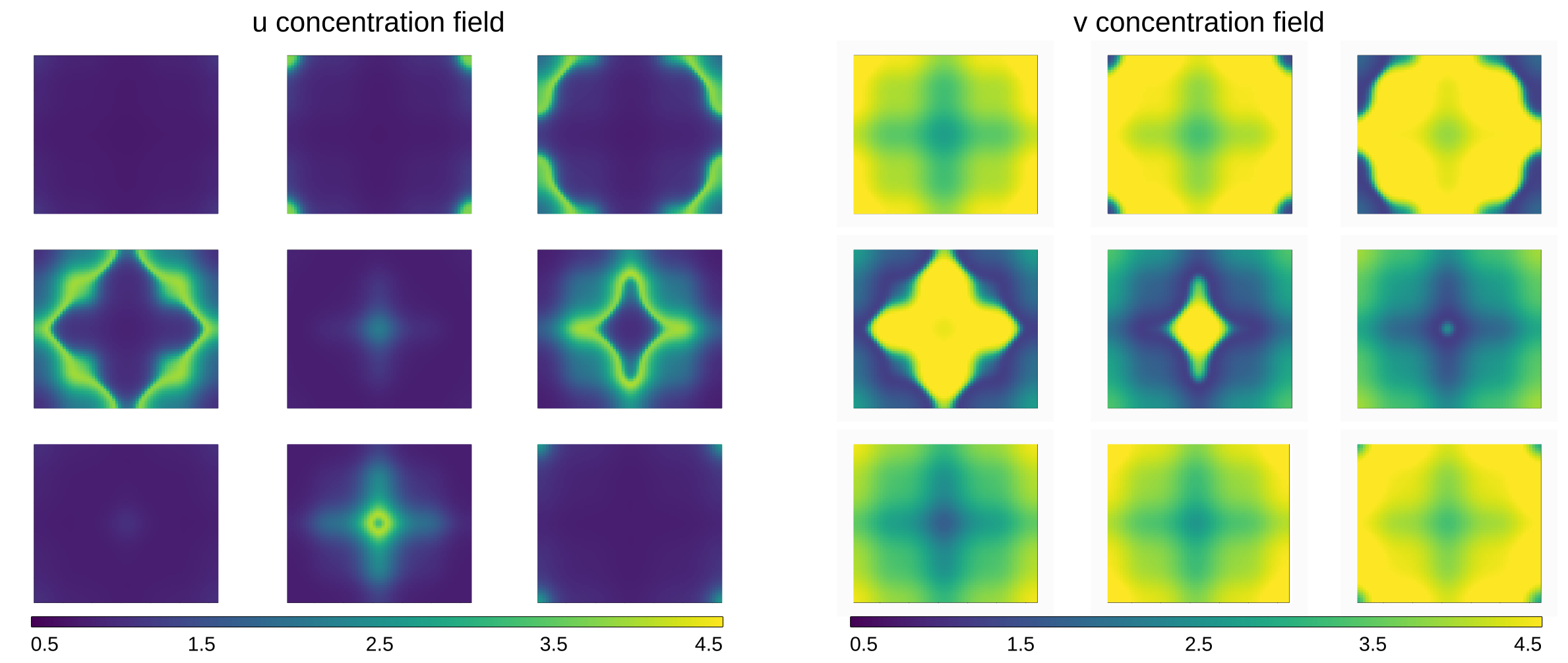}}
\caption{\textbf{Brusselator Dynamics: Concentration Fields $u$ and $v$.} We
show the dynamics of the concentration fields $u(x,t)$ and $v(x,t)$ of the
reaction-diffusion PDEs in equation~\ref{pde_rd_1} with initial conditions 
for $\alpha=0.0$ in equation~\ref{pde_rd_initial}.  Shown is the evolution
of the concentration fields $u,v$ from left to right starting at
time $t=60\Delta{t}$ to time $t=140\Delta{t}$.
After a transient, the concentration field 
dynamics approach a limit cycle having approximately
periodic dynamics.  
}
\label{fig_brusselator_uv_evolve}
\end{figure}

We show how our GD-VAEs can be used to learn features for representing 
the states and dynamic evolution of spatially extended reaction-diffusion systems,
see Figure~\ref{fig_brusselator_uv_evolve}.
Consider the system
\begin{eqnarray}
\label{pde_rd_1}
\frac{\partial u}{\partial t} = D_1 \Delta u + f(u,v),
\hspace{1cm} 
\frac{\partial v}{\partial t} = D_2 \Delta v + g(u,v).
\end{eqnarray}
The $u=u(x,t)$ and $v(x,t)$ give the spatially distributed concentration of
each chemical species at time $t$ with $x \in \mathbb{R}^2$.  We consider the
case with periodic boundary conditions with $x \in [0,L]\times[0,L]$.  We
develop learning methods for investigating the Brusselator system
~\citep{Prigogine_Brussalator_1967,Prigogine_Brussalator_1968}, which is known
to have regimes exhibiting limit
cycles~\citep{Strogatz2018,HirschSmale1974,HirschSmaleMorris2012}.  This
indicates after an initial transient, the orbit of the dynamics
will localize and approach a subset of states topologically 
similar to a circle.  We show how GD-VAE can utilize this
topological information to construct latent spaces for encoding states of the
system.  The Brusselator~\citep{Prigogine_Brussalator_1967,Prigogine_Brussalator_1968} 
has reactions with $f(u,v) = a - (1+b)u+vu^2$ and $g(u,v) = bu -vu^2$.  We take
throughout the diffusivity $D_1=1, D_2=0.1$ and reaction rates $a=1, b=3$.  

We consider initial conditions for the concentration fields $u$ and $v$
parameterized by $\alpha \in [0,1]$ given by 
\begin{eqnarray}
\label{pde_rd_initial}
u(x,y) & = & \alpha \sin(2 \pi x / L_x) + (1-\alpha) \cos^3(2 \pi x / L_x) \\
v(x,y) & = & \alpha \sin(2 \pi y / L_y) + (1-\alpha) \cos^3(2 \pi y / L_y).
\end{eqnarray}
We remark that while the initial conditions are given by a linear combination
of functions the resulting solutions do not satisfy this relation given the
non-linear dynamics of the reaction-diffusion system.  The chemical fields
evolve under periodic boundary conditions within a box length $L_x = L_y = 64$.
We use second order Central-Differences to estimate the spatial derivatives and
Backward-Euler method for the temporal evolution~\citep{Iserles2009,Burden2010}
(py-pde python package~\citep{Zwicker_PyPDE_2020} is used for numerical
calculations). For the discretization, we space the grid points with $\Delta{x}
= 1.0$ and use a time-step of $\Delta{t} = 10^{-3}$.

\begin{figure}[ht]
\centerline{\includegraphics[width=0.99\columnwidth]{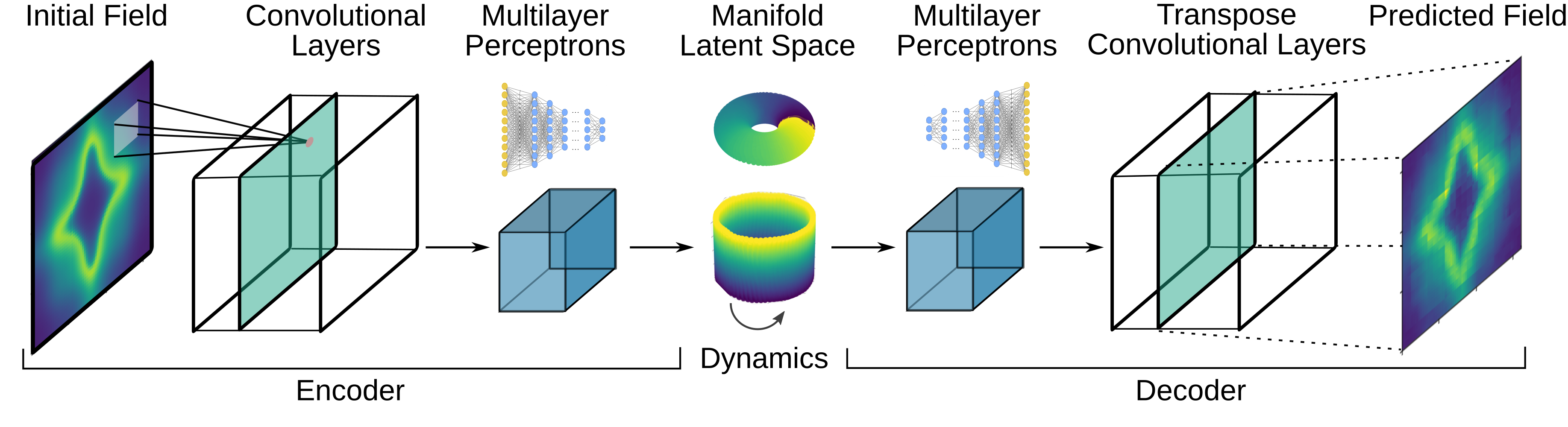}}
\caption{\textbf{GD-VAE Convolutional Neural Network Architecture.} For
spatially extended fields $u,v$, we use architectures based on Convolutional
Neural Networks (CNNs) consisting of the following processing steps.  For the encoder, we
extract features from the spatial fields using Convolutional Neural Networks
(CNNs) which are fed into dense Multilayer Perceptrons (MLPs).  The encoder
serves to map the spatial fields $u,v$ to codes $\mb{z}$ in a manifold latent
space having prescribed topology using our methods discussed in
Section~\ref{sec_manifold_latent_spaces}.  For the decoder, we use MLPs that
feed into Transpose Convolutional Neural Networks (T-CNNs) that serve to
reconstruct from $\mb{z}$ the spatial fields $u,v$ 
We learn the encoders and
decoders using our GD-VAE training framework discussed in
Section~\ref{sec_gd_vae}.
}
\label{fig_brusselator_cnn_dnn}
\end{figure}

\begin{figure}[ht]
\begin{center}
\includegraphics[width=0.8\columnwidth]{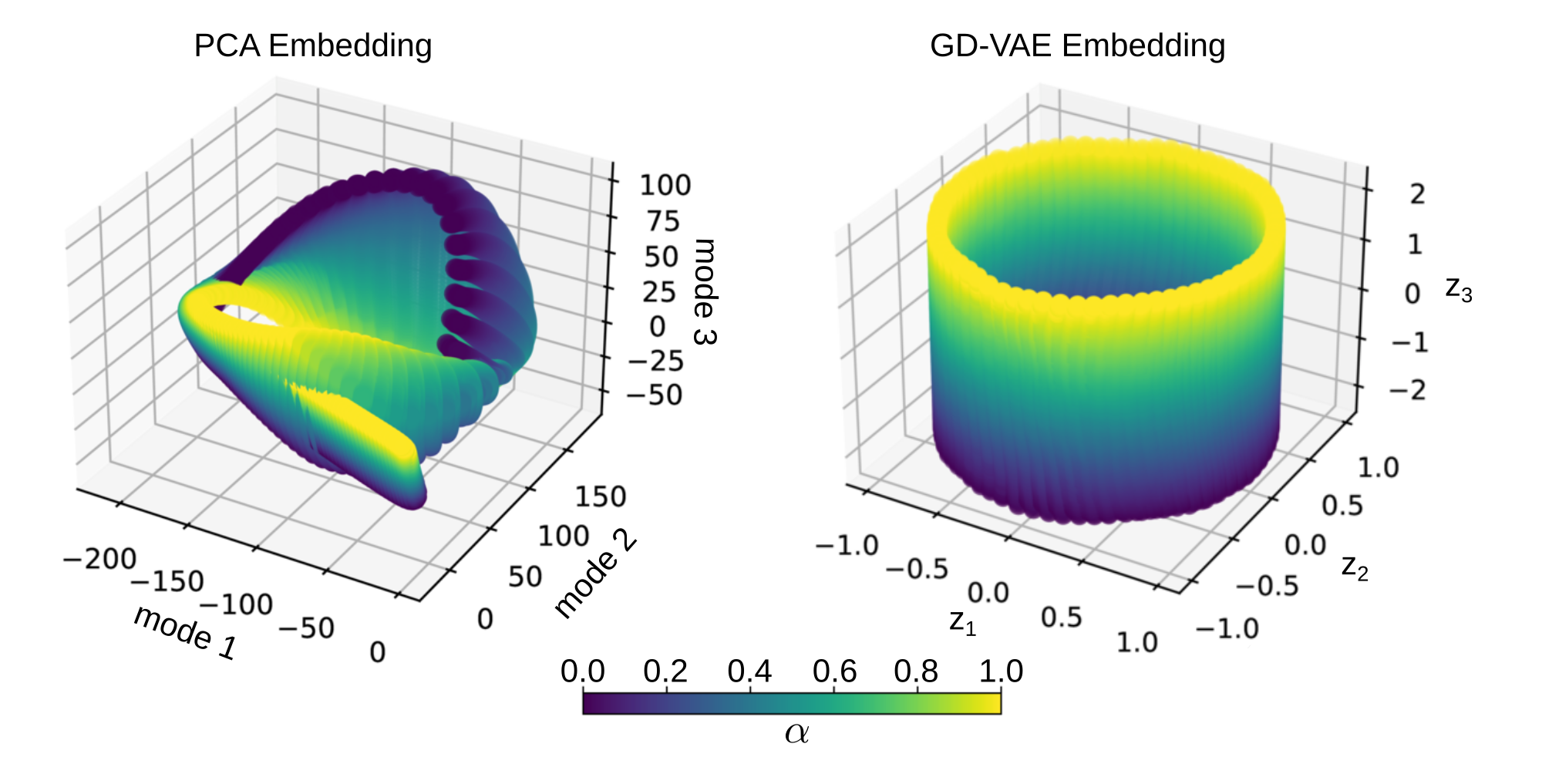}
\end{center}
\caption{\textbf{Embeddings for Brusselator Reaction-Diffusion Dynamics.} 
A Principle Component Analysis (PCA) embedding is compared to GD-VAE embeddings
for solutions of the PDEs in equation~\ref{pde_rd_1}.  While the PCA embedding
provides some information on the transient and limit cycle behaviors, the
latent space representation entangles the temporal and state information,
\textit{(left)}.  The GD-VAE allows for learning a more organized and
parsemoneous embedding for the dynamics where state information is captured by
$(z_1,z_2)$ and time by $z_3$, \textit{(right)}.
}
\label{fig_embeddings_PCA_vs_GD_VAE}
\end{figure}

We show how our GD-VAE methods can be used for solutions to learn a parsimonious
disentangled representations in a manifold latent space.  The dynamics consists
of a brief transient followed by close approach to a limit cycle.  For the
dynamics after the transient, we encode the temporal and state information
using a cylindrical topology.  While a preliminary application of Principle 
Component Analysis (PCA) in a sufficiently large number of dimensions 
can be used to identify the transient and the period of the limit cycle,
it does not provide a well-organized or parsimonious representation of the 
dynamics.  Using from PCA the top three singular vectors, we show the 
entangled embedding in Figure~\ref{fig_embeddings_PCA_vs_GD_VAE}.

We use GD-VAE methods to learn a more organized encoding for representing the
states and system dynamics.  For this purpose, we develop encoders based on
Convolutional Neural Networks (CNNs) and Multilayer Perceptrons (MLPs) to map
from spatial concentration fields $u,v$ to latent codes $\mb{z}$.  The encoder
serves to map the spatial fields $u,v$ to codes $\mb{z}$ in a manifold latent
space having prescribed topology using our methods discussed in
Section~\ref{sec_manifold_latent_spaces}.  We develop decoders based on using
MLPs that control Transpose Convolutional Neural Networks (T-CNNs) to map from
latent codes $\mb{z}$ to construct spatial concentration fields $u,v$, see
Figure~\ref{fig_brusselator_cnn_dnn}.  

We specify a manifold latent space having the geometry of a cylinder
$\mathcal{M} = \mathcal{B}\times \mathbb{R}$ with $\mathcal{B} = S^1$ and axis
in the $z_3$-direction.  We prescribe on this latent space the dynamics having
the rotational evolution 
\begin{equation}
  \mb{z}(t+\Delta t) = 
\begin{bmatrix} 
\cos(\omega \Delta t) & -\sin(\omega \Delta t) & 0 \\ 
\sin(\omega \Delta t) & \cos(\omega \Delta t) & 0 \\ 
0 & 0 & 1  
\end{bmatrix} \mb{z}(t).
\end{equation}
This is expressed in terms of an embedding in $\mathbb{R}^3$. The $\omega$
gives the angular velocity.  This serves to regularize how the encoding of the
reaction-diffusion system is organized in latent space.  

The GD-VAE then is tasked with learning the encoding and decoding mappings for
the reaction-diffusion concentration fields $u,v$ to representations that are
well-organized for capturing and predicting the reaction-diffusion system
states and dynamics.  This organization allows for making multi-step
predictions from the latent space.  Throughout our empirical studies, we use
the angular velocity $\omega = 0.282$ determined from the preliminary PCA
analysis.  In principle, such parameters also can be learned as part of the
training of the GD-VAE.  

For training, we use an architecture with CNNs having four layers
each of which has  (in-channels, out-channels, kernel-size, stride, padding)
with parameters as follows respectively (2,10,3,3,1), \\
(10,20,3,3,1),(20,40,2,2,1),(40,100,5,1,0).  Each layer was followed by a
$ReLU$ activation, except for the last layer which is fed into the MLPs.  The
MLPs have an architecture with layer sizes 100-(latent-space-embed) where
latent-space-embed=3.  The \textit{g-projection} is then applied to the output
of the MLPs to map to the manifold latent space $\mathcal{M}$ using our methods
discussed in Section~\ref{sec_manifold_latent_spaces}.  For the decoder, we use
MLPs with layer sizes (latent-space-embed)-100.  We use T-CNNs having four
layers each of which has (in-channels, out-channels, kernel-size, stride,
padding) with parameters (100, 40, 5, 1, 0), (40, 20, 2, 2, 1), (20, 10, 3, 3,
1), (10, 2, 3, 3, 1).  All layers have a bias and $ReLU$ activations except for
the last layer.

\begin{figure}[ht]
\centerline{\includegraphics[width=0.9\columnwidth]{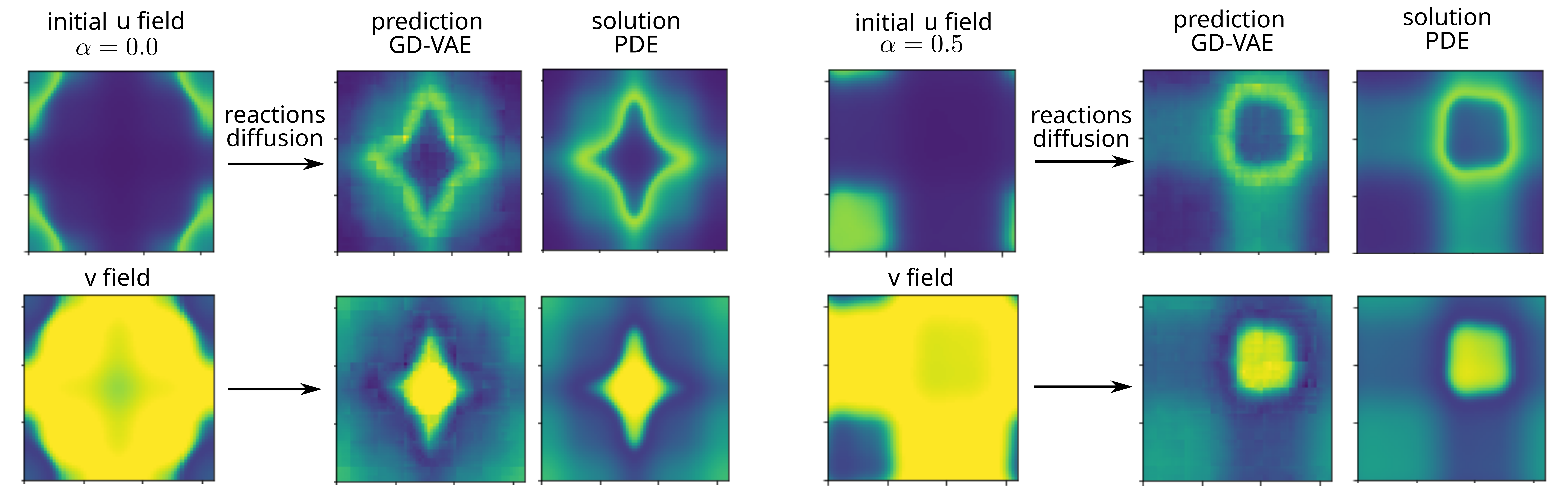}}
\caption{\textbf{Brusselator: GD-VAE Prediction of Concentration Fields $u$ and
$v$.} We show initial concentrations of two chemical species with spatial
distribution given by the fields $u$ and $v$ \textit{(left)}.  We consider how
these evolve after undergoing reaction and diffusion processes according to the
PDEs in equation~\ref{pde_rd_1}.  We show the GD-VAE predictions for the
spatial fields $u, v$ with those of the numerical solutions of the PDEs
\textit{(right)}.  We show the cases for $\alpha=0.0$ and $\alpha=0.5$ starting
at $t=15s$ and predicting the concentration fields $u,v$ at time $t=17s$.  The
$L^1$-relative errors of predictions are given in
Table~\ref{table_brusselator_L1_accur}.
}
\label{fig_brusselator_uv_pred}
\end{figure}

We use our GD-VAE methods for learning representations and for predicting the
evolution dynamics of the reaction-diffusion system.  We 
show a few predictions of the concentration fields by the GD-VAEs in
comparison to the numerical solutions of the PDE in
Figure~\ref{fig_brusselator_uv_pred}.  We characterize the accuracy of the
predictions and reconstructions using $L^1$-relative errors.  We report results
of our methods for multi-step predictions in
Table~\ref{table_brusselator_L1_accur}.  

\begin{table}[ht]
\begin{center}
{
\fontsize{7.3}{10}\selectfont
\selectfont
\begin{tabular}{l|c|ccccc}
\hline 
\rowcolor{atz_table1} 
\textbf{Method} & \textbf{Dim} & \textbf{0.00s} & \textbf{2.00s} & \textbf{4.00s} & \textbf{6.00s} & \textbf{8.00s}  \\ 
\cline{1-7}
GD-VAE & 3 & $3.16e\sminus 02 \pm 5.4e\sminus 03$ & $2.87e\sminus 02 \pm 3.8e\sminus 03$ & $3.63e\sminus 02 \pm 7.0e\sminus 03$ & $3.77e\sminus 02 \pm 6.8e\sminus 03$ & $3.56e\sminus 02 \pm 8.3e\sminus 03$ \\ 
VAE-3D & 3 & $2.61e\sminus 02 \pm 2.9e\sminus 03$ & $2.36e\sminus 02 \pm 2.4e\sminus 03$ & $2.08e\sminus 01 \pm 1.8e\sminus 01$ & $2.61e\sminus 01 \pm 2.3e\sminus 01$ & $2.16e\sminus 01 \pm 1.9e\sminus 01$ \\ 
AE (g-projection) & 3 & $2.96e\sminus 02 \pm 3.5e\sminus 03$ & $2.75e\sminus 02 \pm 2.8e\sminus 03$ & $3.91e\sminus 02 \pm 9.7e\sminus 03$ & $4.25e\sminus 02 \pm 1.2e\sminus 02$ & $2.99e\sminus 02 \pm 2.5e\sminus 03$ \\ 
AE (no projection) & 3 & $2.36e\sminus 02 \pm 1.8e\sminus 03$ & $2.19e\sminus 02 \pm 1.5e\sminus 03$ & $3.49e\sminus 01 \pm 3.2e\sminus 01$ & $1.88e\sminus 01 \pm 1.6e\sminus 01$ & $1.55e\sminus 01 \pm 1.3e\sminus 01$ \\
\hline
\end{tabular} \\
}
\end{center}
\caption{\textbf{Brusselator: $L^1$-Prediction Accuracy.} The reconstruction 
$L^1$-relative errors and standard error over 5 training
trials.  We compare the GD-VAE
approach with other methods for learning latent space representations for the
reaction-diffusion dynamics.  We consider the relative error of
predictions when receiving the concentration field at time $t_1$ and making
a prediction of the concentration field at time $t_1 + \tau$.  We investigate
the multi-step predictions up to time-scale $\tau=8.00s$.  This involves a
composition of the single-step predictions over multiple steps for dynamics
that has gone entirely through at least one period of the limit cycle.  We see
the geometric structure in the latent space greatly enhances the stability and
resulting accuracy of predictions. 
}
\label{table_brusselator_L1_accur}
\end{table}

We find that GD-VAEs are able to learn representations in the manifold latent
spaces capable of making good predictions of the dynamics.  The manifold latent
space is particularly advantageous for multi-step predictions by helping to
constrain the encodings promoting more robust learning over a lower dimensional
space and smaller subset of points. In this geometric setting, learning only
needs to be performed over the subset of points on the manifold.  This enhances 
the statistical power of the training data.  This also simplifies the type of
encoder and decoder response functions that need to be learned relative to
the higher dimensional setting of $\mathbb{R}^n$.  In addition, during
multi-step predictions we see the geometric constraints also enhance stability.
This arises from the latent space dynamics always being confined within the
manifold.  As a result, it is less likely for the one-step updates to map to a
code $\mb{z}$ in unfamiliar parts of the latent space away from those locations
characterized during training.
The lower dimension of the geometric
latent space also provides similar benefits with encoding.  This arises from 
points being more concentrated and providing more statistical power 
in learning a model for the local
responses of the underlying reaction-diffusion system.  The more organized
representations are also more amenable to interpretation.  

These results indicate some of the ways our GD-VAE approaches can be used to leverage
qualitative information about the dynamics both to enhance learning and to gain
insights into system dynamics.  The methods provide practical ways to learn
parsimonious representations capable of providing quantitatively accurate
predictions for high dimensional dynamical systems having latent 
geometric structures.  The GD-VAEs can be used to obtain representations for 
diverse learning tasks for many types of dynamical systems.

\section*{Software Package}
We have developed software packages for the introduced GD-VAE methods.  This
includes python implementations of our geometric projections (g-projection),
manifold latent space representations, custom back-propagation approaches for 
modular training, and examples.  The python package can be installed 
using \verb+pip install gd-vae-pytorch+ or 
downloaded from our website.  For more 
information, see \url{https://atzberger.org/}.

\section*{Conclusions}
We introduced GD-VAEs for learning representations of nonlinear 
dynamics on manifold latent spaces having general topologies 
and geometries.  The methods allow for  
learning representations with prescribed geometric properties 
which can be used to help enhance
the robustness of dynamical predictions, yield more interpretable results, or
provide further insights into the underlying mechanisms generating observed
behaviors. The methods also allow for leveraging qualitative information
and analysis of dynamical systems for use in data-driven learning. 
We performed several benchmark studies to validate and characterize
the methods.  This included making comparisons of GD-VAEs with POD,
DMD, and more conventional AEs.  Our results indicate how the 
non-linear approximation properties of neural networks combined
with geometric inductive biases can be used to help 
improve the reductions of representations and the accuracy of 
predictions and reconstructions. We also presented 
results for constrained mechanical systems, and 
the non-linear dynamics of parameterized PDEs, 
Burgers' equations, and reaction-diffusion systems.
The results indicate some of the ways  
geometric and topological information
present opportunities to simplify model representations, 
aid in interpretability, and enhance 
robustness of predictions.  The GD-VAEs can be 
used to obtain representations for diverse types 
of learning tasks involving dynamics.

\section*{Acknowledgements} 
Authors research supported by grants DOE Grant ASCR PHILMS DE-SC0019246, NSF
Grant DMS-1616353, and NSF-DMS-2306101.  Also to R.N.L.  support by a donor to
UCSB CCS SURF program.  Authors also acknowledge UCSB Center for Scientific
Computing NSF MRSEC (DMR1121053) and UCSB MRL NSF CNS-1725797.  P.J.A. would
also like to acknowledge a hardware grant from Nvidia.

\appendix

\newpage
\clearpage

\section*{Appendix}

\section{Backpropagation of Encoders for Non-Euclidean Latent Spaces given by
General Manifolds}
\label{sec_backprop}

\begin{figure}[ht]
\centerline{\includegraphics[width=0.8\columnwidth]{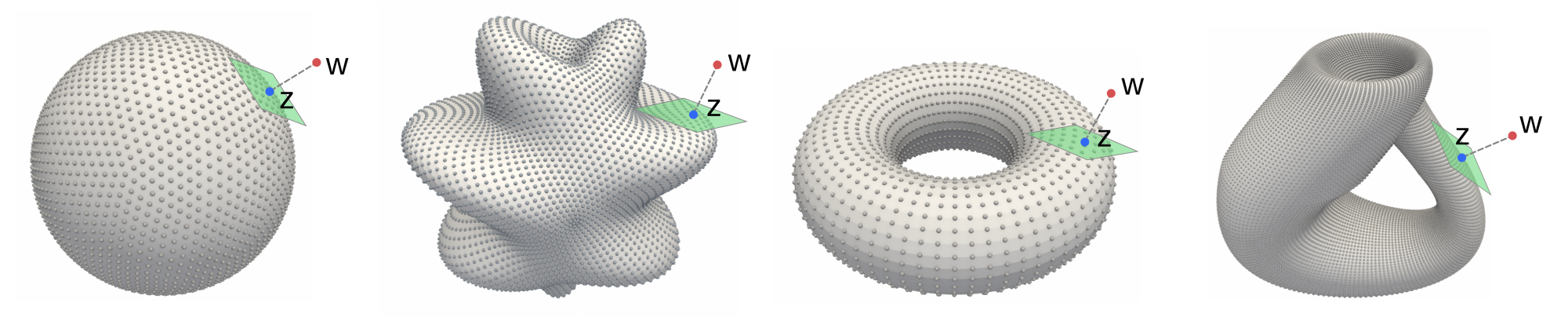}}
\caption{\textbf{Learnable Mappings to Manifold Surfaces.}  We develop methods
based on point cloud representations embedded in $\mathbb{R}^{n}$ for
learning latent manifold representations having general geometries and
topologies. 
}
\label{fig:manifold_map}
\end{figure}

We develop methods for using back-propagation to learn encoder maps from
$\mathbb{R}^d$ to general manifolds $\mathcal{M}$.  We perform learning using
the family of manifold encoder maps of the form $\mathcal{E}_\theta =
\Lambda(\tilde{\mathcal{E}}_\theta(x))$.  This allows for use of latent spaces
having general topologies and geometries.  We represent the manifold as an
embedding $\mathcal{M} \subset \mathbb{R}^{2m}$ and computationally use
point-cloud representations along with local gradient information, see
Figure~\ref{fig:manifold_map} and~\ref{fig:manifold_map_embed}.  
To allow for $\mathcal{E}_\theta$ to be
learnable, we develop approaches for incorporating our maps into general
back-propagation frameworks.

For a manifold $\mathcal{M}$ of dimension $m$, we can represent it by an
embedding within $\mathbb{R}^{2m}$, as supported by the Whitney Embedding
Theorem~\cite{Whitney1944}.  We let $\mb{z} = \Lambda(\mb{w})$ be a mapping
$\mb{w} \in \mathbb{R}^{2m}$ to points on the manifold $\mb{z} \in
\mathcal{M}$.  This allows for learning within the family of manifold encoders
$w = \tilde{\mathcal{E}}_\theta(x)$ any function from $\mathbb{R}^d$ to
$\mathbb{R}^{2m}$.  This facilitates use of deep neural networks and other
function classes.  In practice, we shall take $\mb{z} = \Lambda(\mb{w})$ to map
to the nearest location on the manifold.  We can express this as the
optimization problem $$ z^* = \arg\min_{z \in \mathcal{M}} \frac{1}{2}\|w -
z\|_2^2.  $$ We can always express a smooth manifold using local coordinate
charts $\sigma^k(u)$, for example, by using a local Monge-Gauge quadratic fit
to the point cloud~\citep{Atzberger_GMLS_Surf_PDE_2019}.  We can express $z^* =
\sigma^k(u^*)$ for some chart $k^*$.  In terms of the coordinate charts
$\{\mathcal{U}_k\}$ and local parameterizations $\{\sigma^{k}(u)\}$ we can
express this as $$ u^*,k^* = \arg\min_{k,u \in \mathcal{U}_k} \frac{1}{2}\|w -
\sigma^{k}(u)\|_2^2, $$ where $\Phi_k(u,w) = \frac{1}{2}\|w -
\sigma^k(u)\|_2^2$.  The $w$ is the input and $u^*,k^*$ is the solution sought.
For smooth parameterizations, the optimal solution satisfies $$G = \nabla_z
\Phi_{k^*}(u^*,w) = 0.$$  During learning we need gradients $\nabla_w
\Lambda(w) = \nabla_w z$ when $w$ is varied characterizing variations of points
on the manifold $z = \Lambda(w)$.  We derive these expressions by considering
variations $w = w(\gamma)$ for a scalar parameter $\gamma$.  We can obtain the
needed gradients by determining the variations of $u^* = u^*(\gamma)$.  We can
express these gradients using the Implicit Function Theorem as 
$$
0 = \frac{d}{d\gamma} 
G(u^*(\gamma),w(\gamma))
= \nabla_u G \frac{du^*}{d\gamma} 
+ 
\nabla_w G \frac{dw}{d\gamma}.
$$
This implies 
$$
\frac{du^*}{d\gamma} = -\left[\nabla_u G \right]^{-1}
\nabla_w G \frac{dw}{d\gamma}.
$$
As long as we can evaluate at $u$ these local gradients $\nabla_u G$, $\nabla_w
G$, $dw/d\gamma$, we only need to determine computationally the solution $u^*$.
For the back-propagation framework, we use these to assemble the needed
gradients for our manifold encoder maps $\mathcal{E}_\theta =
\Lambda(\tilde{\mathcal{E}}_\theta(x))$ as follows.

We first find numerically the closest point in the manifold $z^* \in
\mathcal{M}$ and represent it as $z^* = \sigma(u^*) = \sigma^{k^*}(u^*)$ for
some chart $k^*$.  In this chart, the gradients can be expressed as
$$
G = \nabla_u \Phi(u,w) = -(w - \sigma(u))^T\nabla_u \sigma(u).
$$
We take here a column vector convention with
$\nabla_u \sigma(u) = [\sigma_{u_1} | \ldots | \sigma_{u_k}]$.  We next compute 
$$
\nabla_{u} G = \nabla_{uu} \Phi = \nabla_u\sigma^T \nabla_u \sigma - (w -
\sigma(u))^T\nabla_{uu} \sigma(u)
$$
and 
$$
\nabla_w G = \nabla_{w,u} \Phi = -I \nabla_{u} \sigma(u).
$$
For implementation it is useful to express this in more detail component-wise as
$$
[G]_i = -\sum_k (w_k - \sigma_k(u))\partial_{u_i} \sigma_k(u),
$$
with 
\begin{eqnarray}
\nonumber [\nabla_u G]_{i,j} & = & [\nabla_{uu} \Phi]_{i,j} = \sum_k
\partial_{u_j} \sigma_k(u)\partial_{u_i} \sigma_k(u) \\ \nonumber & - &
\sum_k (w_k - \sigma_k(u))\partial_{u_i,u_j}^2 \sigma_k(u) \\ \nonumber
[\nabla_w G]_{i,j} & = & [\nabla_{w,u} \Phi]_{i,j} \\ \nonumber & = & -\sum_k
\partial_{w_j} w_k\partial_{u_i} \sigma_k(u) = -\partial_{u_i} \sigma_j(u).
\end{eqnarray}
The final gradient is given by 
$$
\frac{d\Lambda(w)}{d\gamma}
= \frac{dz^*}{d\gamma} = 
\nabla_u \sigma \frac{du^*}{d\gamma} 
= -\nabla_u \sigma\left[\nabla_u G \right]^{-1}
\nabla_w G \frac{dw}{d\gamma}.
$$

In summary, once we determine the point $z^* = \Lambda(w)$ we need only
evaluate the above expressions to obtain the needed gradient for learning via
back-propagation
$$
\nabla_\theta \mathcal{E}_\theta (x) = 
\nabla_w \Lambda(w) \nabla_\theta \tilde{\mathcal{E}}_\theta(x), \; w = \tilde{\mathcal{E}}_\theta(x).
$$
The $\nabla_w \Lambda$ is determined by ${d\Lambda(w)}/{d\gamma}$ using $\gamma
= w_1, \ldots w_n$.  In practice, the $\tilde{\mathcal{E}}_\theta(x)$ is
represented by a deep neural network from $\mathbb{R}^d$ to $\mathbb{R}^{2m}$.
In this way, we can learn general encoder mappings $\mathcal{E}_\theta (x)$
from $x \in \mathbb{R}^d$ to general manifolds $\mathcal{M}$.

\section{Role of Latent Space Geometry in Training}
\label{sec_role_geo_struc}

\begin{figure}[h]
\centerline{
\includegraphics[width=0.75\columnwidth]{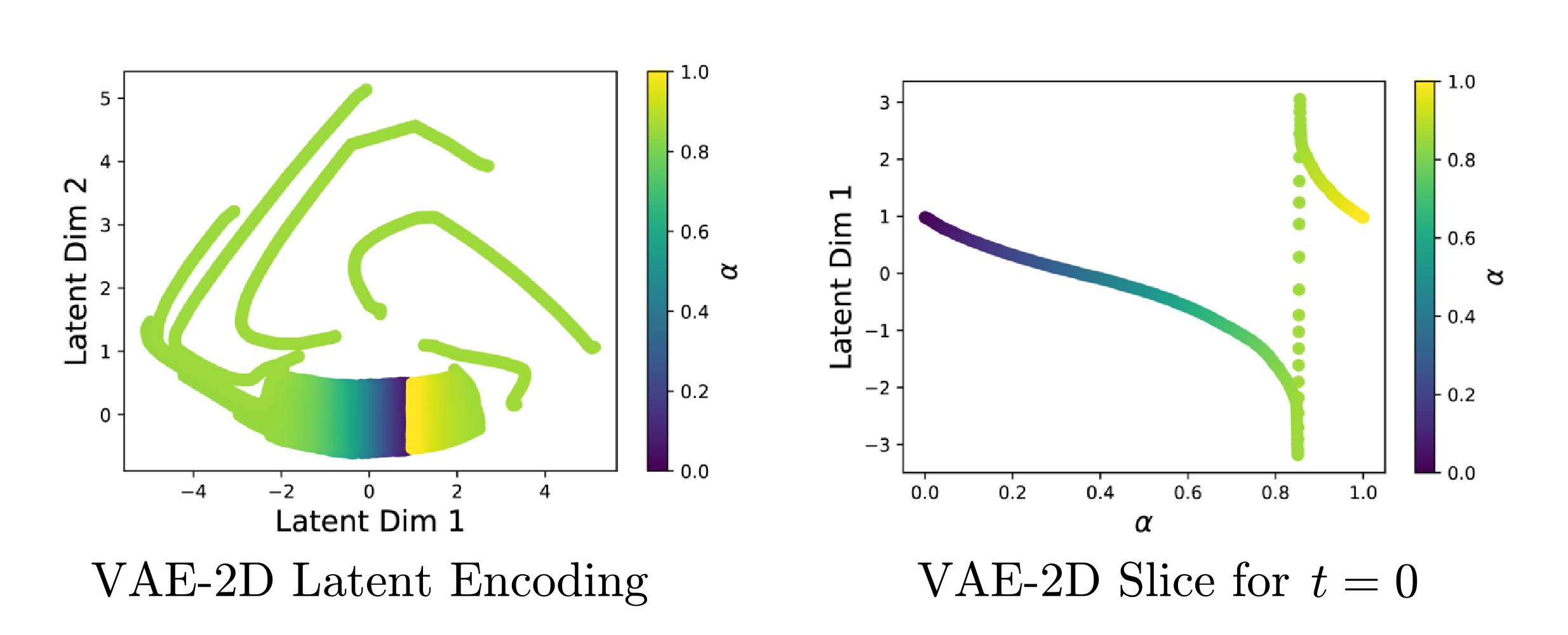}    
}
\caption{\textbf{Latent Space with Incompatible Geometry}  We show a typical
behavior encountered during training for dynamical systems having periodic
parameterization $\alpha \in [0,1]$ when using a latent space representation
based on $\mathbb{R}^2$.  While for a finite duration of time $t \in [0,T]$
there are in principle embeddings for such dynamics into $\mathbb{R}^2$ as in
equation~\ref{equ_rep_cone_proj}, these can be difficult to find during
training, and further not possible to disentangle time and state configurations 
into separate dimensions.
From the periodicity of inputs and the continuity of the encoder a
typical behavior instead is for most states to be mapped to a patch and for the
mapping to incur a penalty by rapidly loop back over only a small subset of
inputs, here for responses for $\alpha\sim 0.85$.  The continuous function
approximates what would have been a discontinuity \textit{(right)}.  This rapid
variation of the encoder results in states being mapped to a scattered set of
codes which provide a poor representation for the dynamics.  This is exhibited
by the long spirals that show segments of the dynamics over time and how the
states are mapped to the latent codes for $\alpha \sim 0.85$, \textit{(left)}.
Our GD-VAEs provide approaches to avoid this issue by allowing for general
manifold latent spaces which maintain low dimension while accommodating such
topology.  }
\label{fig_topo_incompatible}
\end{figure}
In practice, the encodings that organize the data in a manner convenient for
representing the dynamic evolution can be challenging to learn in data-driven
methods.  In cases where the embedding space is too low dimensional or
incompatible with the intrinsic topology this is further compounded.  To
illustrate some of the issues that arise, we consider a dynamical system with a
periodic parameterization using as our example the Burgers' equation in
equation~\ref{equ_burgers_equ} discussed in Section~\ref{sec_burgers_equ}.  For
this data set, it is natural to consider representing the system state and the
dynamics using the geometry of a cylinder.  In principle, it is possible 
to obtain a representation of the dynamics if restricted over a finite
duration of time using only a two dimensional Euclidean space $\mathbb{R}^2$
as in equation~\ref{equ_rep_cone_proj}.  This is similar to a projection 
of a cone-shaped manifold in three dimensions projected to the 
two dimensional plane.  

However, in practice this can be difficult to
work with since there is an inherent tension during training between the
periodicity and the natural ways the typical encoders will operate.  
This arises from the continuity of the encoder model class used in training
that maps from the dynamical system solutions to the $\mathbb{R}^2$ latent 
space.  We find that training proceeds initially by mapping a collection
of states to a smooth patch of codes in the latent space.  As a 
consequence of continuity, the encoder for the states associated with
system responses for $\alpha$ and $\alpha'\equiv \alpha \pm 1.0\ \textrm{mod}\ 1.0$ 
need to map to similar codes.  This results in an encoder that exhibits for a subset of 
states an extreme sensitivity to inputs that results in rapidly varying 
the code to loop back to accommodate the periodicity, see $\alpha = \lim \uparrow 0.85$ and 
$\alpha' = \lim \downarrow 0.85$ in Figure~\ref{fig_topo_incompatible}.  
Again, the issue arises since the states must map continuously to the latent space for
all encoders encountered during training.  Unless the map happens already to form a 
circle or other periodic structure in $\mathbb{R}^2$ there will be a subset of 
points that arise with rapid variation.  This results in a subset
of points having a poor encoding with states mapped to codes scattered in the
latent space.  This provides a poor basis for representing and predicting the
system dynamics.  When encoding over time we see disorganized fragments,
connected again by transitions having rapid variations.  
We show this typical behavior observed during such training in
Figure~\ref{fig_topo_incompatible}.  This indicates the importance
of choosing a latent space with either a sufficiently large
number of dimensions or which has a topology that is compatible with
the data set.  Our GD-VAE approach allows for keeping the 
latent space dimensionality small by allowing for accommodating general 
topologies in the latent space using our methods in 
Section~\ref{sec_manifold_latent_spaces}.

\section{Identifying Latent Geometric Structures using VAE Covariance $\bsy{\Sigma}_e$}
\label{sec_extract_geo_from_var}

The VAE training can be performed with a learnable covariance structure $\bsy{\Sigma}_e$
with entries $[\bsy{\Sigma}_e]_{ij} = \Sigma_{e,ij}$, see equation~\ref{equ_vae}.  In VAE the
$KL$-divergence regularization term is typically used to drive the encoding
points to match a standard Gaussian distribution or other prior.  The
coding-point distribution arises from a combination of the scattering of the
coding of points from the encoded training data points and from the Gaussian
noise controlled by the variance $\bsy{\Sigma}_e$, see equation~\ref{equ_vae}.
Consider covariance $\bsy{\Sigma}_e = \mbox{diag}(\sigma_{e,i}^2)$, where
$\sigma_{e,i}^2 = \Sigma_{e,ii}$.  If components of $\sigma_{e,i}^2$ are held
fixed to be a small value then the coding-point distribution can only arise
from scattering of the encoding points.  However, if $\sigma_{e,i}^2$ is
learnable, then the encoder can map to a narrow distribution in some of the
coding points with the Gaussian noise with variance $\sigma_{e,i}^2$
compensating to satisfy the $KL$-regularization term. 

\begin{figure}[ht]
\centerline{\includegraphics[width=0.99\columnwidth]{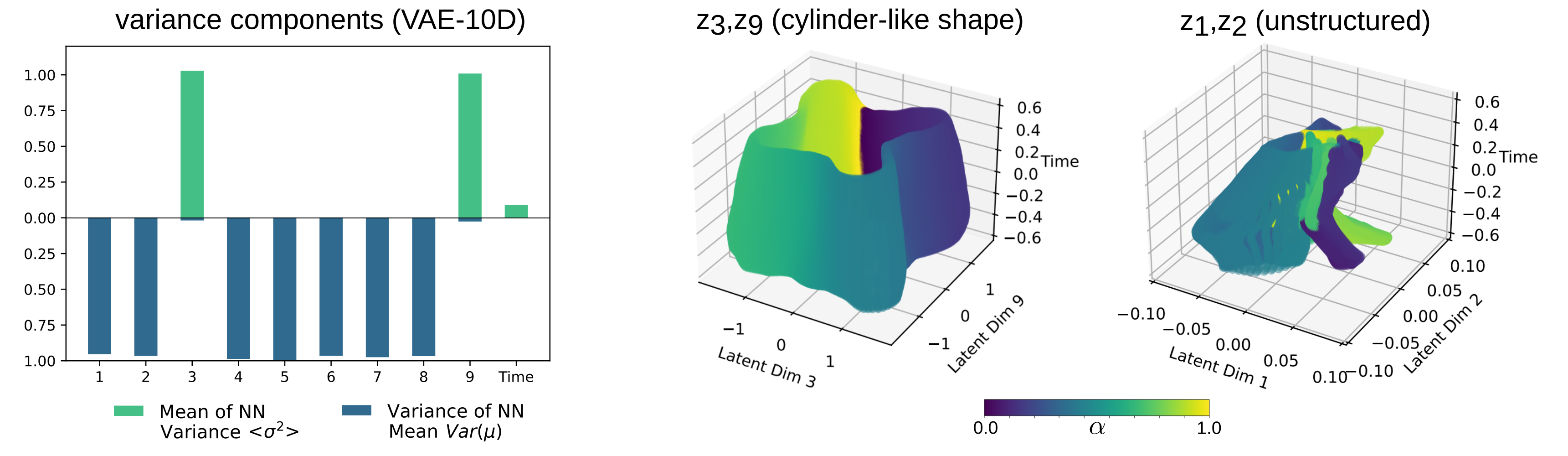}}
\caption{\textbf{Extracting Latent Geometric Structure using Learned VAE
Variances.} We use the $\sigma_{e,i}$ and $\mu_{e,i}$ of VAEs to further
extract geometric information from data and to perform reductions similar
to~\citep{Soljacic2020}. By looking at mini-batches of data, we can compute two
statistics (i) the mean of the variance (mov) $q_{i,mov}$ and (ii) the variance
of the mean (vom) $q_{i,vom}$, see equations~\ref{equ_q_i_mov}
and~\ref{equ_q_i_vom}.  We show results for Burgers' Equation with embedding in
an unconstrained $10$ dimensional space. We see the most relevant feature
dimensions are the ones with $q_{i,vom}$ large and $q_{i,mov}$ small
\textit{(left)}.  This yields a reduced three dimensional embedding with a good
representation for the state information using $z_1,z_9$ coordinates
\textit{(middle)}.  In contrast, using two arbitary dimensions such as
coordinates $z_1,z_2$ yields a poor representation which is entangled and
mapped to small scales \textit{(right)}.  This provides further ways  to
extract geometric information from the data and representations of latent
structures.}
\label{fig:vae_identification}
\end{figure}

We have found in practice in some cases this can be used to identify geometric
structure within the data.  We perform the embedding into a higher dimensional
space than is needed, we find smaller values of $\bsy{\Sigma}_e$ correlate with
encoding to a lower dimensional manifold structure, similar to
~\citep{Soljacic2020}.  We show how this can be used in practice to find
reduced embeddings.  By looking at mini-batches of data with $N_b$ samples, we
can compute two statistics (i) the mean of the variance (mov) 
\begin{eqnarray}
\label{equ_q_i_mov}
q_{i,mov} =
\frac{1}{N_b}\sum_{\ell=1}^{N_b} \sigma_{e,i}^2(\mb{x}^{(\ell)}),
\end{eqnarray}
and (ii) the
variance of the mean (vom) 
\begin{eqnarray}
\label{equ_q_i_vom}
q_{i,vom} = \frac{1}{N_b} \sum_{\ell=1}^{N_b}
\mu_{e,i}^2(\mb{x}^{(\ell)}) - \bar{\mu}_{e,i}^2,
\end{eqnarray}
where $\bar{\mu}_{e,i} =
\frac{1}{N_b} \sum_{\ell=1}^{N_b} \mu_{e,i}(\mb{x}^{(\ell)})$.

When allowing the variance to be trainable, we can compute $q_{i,vom}$ and
$q_{i,mov}$ to help identify important feature directions in latent space.  We
show results for training the Burgers' Equation embedding the cylinder topology
case of Section~\ref{sec_burgers_topo} in Figure~\ref{fig:vae_identification}.
We see important feature components of the encoded data have the characteristic
signature of having a large $q_{i,vom}$ and a small $q_{i,mov}$.  If we reduce
the encoding to just these components, we can obtain a lower dimensional
embedding for the data.  We find this embedding have a cylindrical geometric
structure providing a good latent space representation of the system state
information.  In contrast, if we were to choose any two arbitrary latent code
dimensions, such as $z_1,z_2$ we obtain a poor entangled representation, see
Figure~\ref{fig:vae_identification}.  This allows for using information from
the $\sigma_{e,i}$ and $\mu_{e,i}$ to further extract geometric structure from
data.  This can be used to formulate latent manifold spaces for use with our
GD-VAE approaches.

\newpage
\clearpage
\bibliography{paper_database}

\end{document}